\documentclass[letterpaper, 10 pt, conference]{ieeeconf}  

\IEEEoverridecommandlockouts                              

\usepackage{amsmath} 
\usepackage{amssymb}  
\usepackage[usenames,dvipsnames]{color}
\usepackage{graphicx}
\usepackage{ifthen}
\usepackage{url}
\usepackage{bbm}
\usepackage{multirow}

\usepackage{dsfont}
\usepackage{mathrsfs}

\usepackage{soul}

\makeatletter
\let\NAT@parse\undefined
\makeatother

\usepackage[numbers]{natbib}

\usepackage[normalem]{ulem}

\title{\LARGE \bf
Robobarista: Learning to Manipulate Novel Objects \\ via Deep Multimodal Embedding
}

\author{Jaeyong Sung, Seok Hyun Jin, Ian Lenz, and Ashutosh Saxena
\thanks{Authors are with Department of Computer Science, Cornell University. Email:
{\tt\small \{jysung,sj372,ianlenz,asaxena\} @cs.cornell.edu}
Jaeyong Sung is also a visiting student researcher at Stanford University. Ashutosh Saxena is also with Brain of Things, Inc.}%
}

\newcommand{\argmin}{\arg\!\min}
\newcommand{\argmax}{\arg\!\max}

\newcommand{\header}[1]{\smallskip\noindent\textbf{#1}}

\newcommand{\jae}[1]{\textcolor{Black}{#1}}
\newcommand{\jin}[1]{\textcolor{Black}{#1}}
\newcommand{\ian}[1]{\textcolor{Black}{#1}}

\newboolean{include-notes}
\setboolean{include-notes}{true}
\newcommand{\jaenote}[1]{\ifthenelse{\boolean{include-notes}}%
 {\textcolor{PineGreen}{[\textbf{JAE: #1}]}}{}}
\newcommand{\jinnote}[1]{\ifthenelse{\boolean{include-notes}}%
 {\textcolor{RedViolet}{[\textbf{JIN: #1}]}}{}} 
\newcommand{\ilnote}[1]{\ifthenelse{\boolean{include-notes}}%
 {\textcolor{Bittersweet}{[\textbf{IAN: #1}]}}{}}

\definecolor{darkgreen}{rgb}{0.0, 0.2, 0.13}

\begin{document}

\maketitle
\thispagestyle{empty}
\pagestyle{empty}


\begin{abstract}
There is a large variety of objects and appliances in human environments, such as stoves, 
coffee dispensers, juice extractors, and so on.  It is challenging for a roboticist to program a 
robot for each of these object types and for each of their instantiations.
In this work, we present a novel approach to manipulation planning
based on the idea that many household objects share similarly-operated object parts.
We formulate the manipulation planning as a structured prediction problem
and learn to transfer manipulation strategy across different objects
by embedding point-cloud, natural language, and manipulation trajectory data 
into a shared embedding space using a deep neural network. 
In order to learn semantically meaningful spaces throughout our network, 
we introduce a method for pre-training its lower layers for multimodal feature embedding 
and a method for fine-tuning this embedding space using a loss-based margin.
In order to collect a large number of manipulation demonstrations for different objects,
we develop a new crowd-sourcing platform called Robobarista.
We test our model on our dataset consisting of 116 objects and appliances with 249 parts
along with 250 language instructions, for which there are 1225 crowd-sourced 
manipulation demonstrations.
We further show that our robot with our model can even prepare a cup of a latte
with appliances it has never seen before.\footnote{Parts of this work were presented at ISRR 2015 (\citet{sung_robobarista_2015})}
\end{abstract}



\section{Introduction}

\begin{figure}[t]
  \begin{center}
    \includegraphics[width=\columnwidth,trim={2.8cm 0cm 2.8cm 1.7cm},clip]{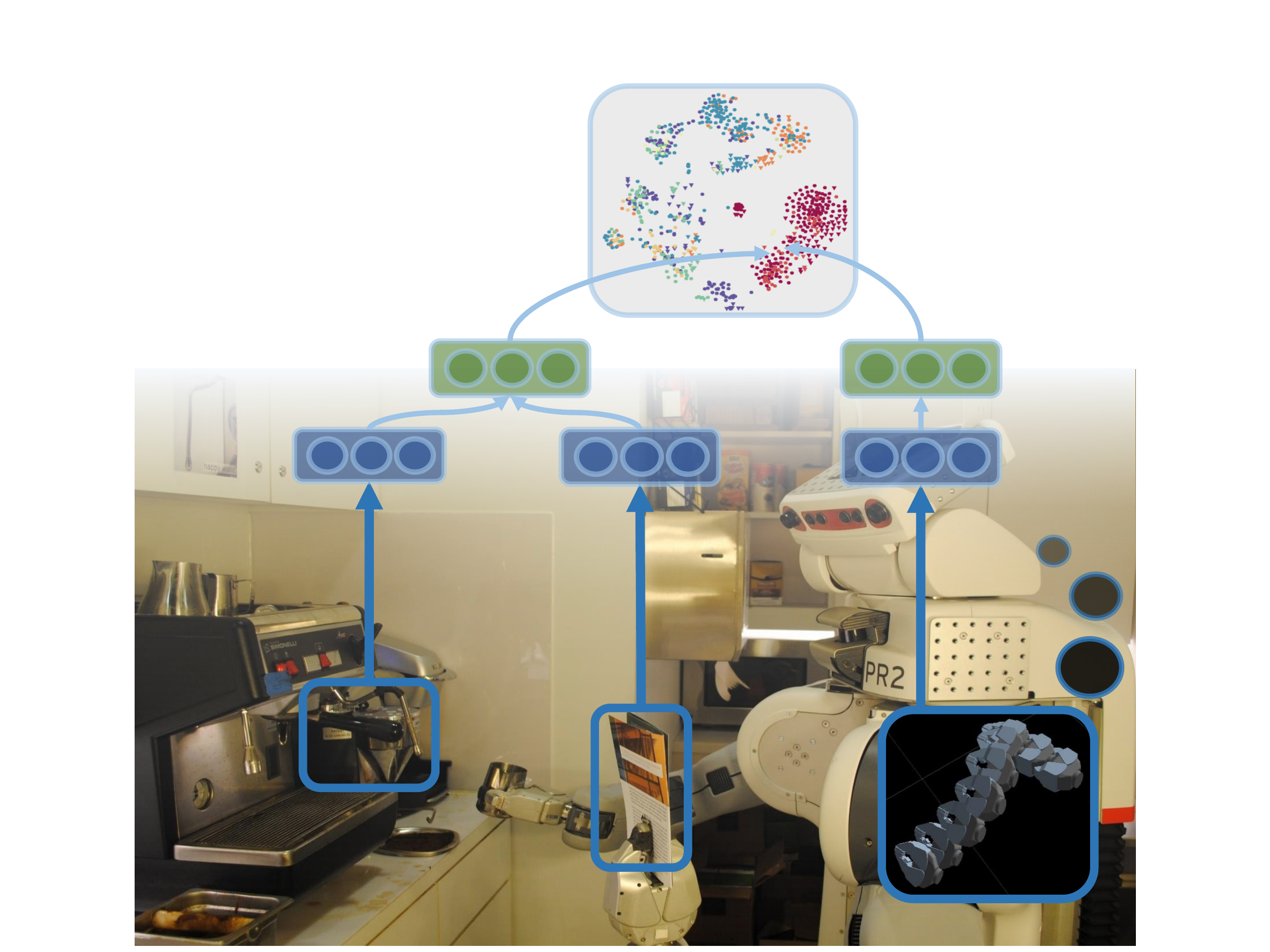}
  \end{center}
  \vskip -.1in
  \caption{ 
  \textbf{First encounter of an espresso machine} by our PR2 robot. Without ever having seen the machine before,  given language instructions and a point-cloud from Kinect sensor, our robot is 
  capable of finding appropriate  manipulation trajectories from prior experience using our deep multimodal embedding model.}
  \label{fig:robobarista_main}
\end{figure}

Consider the espresso machine in Figure~\ref{fig:robobarista_main} ---
even without having seen this machine before, a person can prepare a cup of latte
by visually observing the machine and
reading an instruction manual. 
This is possible because humans have vast prior experience with manipulating
differently-shaped objects that share common parts such as `handles' and `knobs.'
In this work, our goal is to give robots the same capabilites -- specifically,
to enable robots to generalize their manipulation ability to novel objects and tasks (e.g. toaster, sink, water fountain, toilet, soda dispenser, etc.).
\jae{We build an algorithm that uses a large knowledge base of
manipulation demonstrations to infer an appropriate manipulation trajectory for
a given pair of point-cloud and natural language instructions.}

If the robot's sole task is to manipulate one specific espresso machine or 
just a few types of `handles', 
a roboticist could manually program the exact sequence to be executed.
However, human environments contain a huge variety of objects, which makes this approach
un-scalable and infeasible.
Classification of objects or object parts (e.g. `handle') alone does not provide enough information for robots to
actually manipulate them, \ian{since semantically-similar objects or parts might be operated completely differently -- consider,
for example, manipulating the `handle' of a urinal, as opposed to a door `handle'.}
Thus, rather than relying on scene understanding techniques \citep{blaschko2008learning,li2009towards,girshick2011object},
we directly use 3D point-clouds for manipulation planning
using machine learning algorithms.

\begin{figure*}[tb]
\begin{center}
\includegraphics[width=\textwidth,trim={3mm 0 1mm 0}]{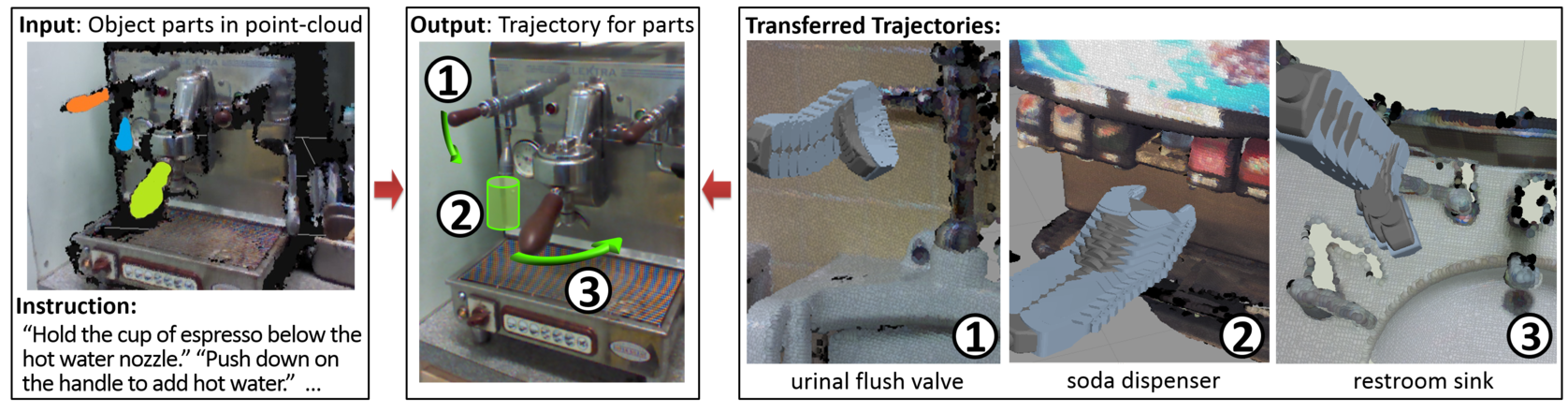}
\end{center}
\caption{\textbf{Mapping object part and natural language instruction input to manipulation trajectory output.} Objects such as the espresso machine
consist of distinct object parts, each of which requires a distinct manipulation trajectory for manipulation. 
For each part of the machine, we can re-use a manipulation trajectory that was used for some other object with 
similar parts. So, for an object part in a point-cloud (each object part colored on left), we 
can find a trajectory used to manipulate some other object (labeled on the right) that can be 
\emph{transferred} (labeled in the center). With this approach, a robot can operate a new and previously 
unobserved object such as the `espresso machine', 
by successfully transferring trajectories from other completely different but previously observed objects.
Note that the input point-cloud is very noisy and incomplete (black represents missing points).}
\label{fig:espresso_transfer}
\end{figure*}

The key idea of our work is that objects designed for use by humans
share many similarly-operated \emph{object parts} such as `handles', `levers', `triggers', and `buttons';
thus, manipulation motions can be transferred even between completely different objects
if we represent these motions  with respect to these parts.
For example, even if the robot has never seen an espresso machine before, it should be able to
manipulate it if it has previously seen similarly-operated parts 
of other objects such as a urinal, soda dispenser, or restroom sink,
as illustrated in Figure~\ref{fig:espresso_transfer}.
Object parts that are operated in similar fashion may not \jae{necessarily} carry the same part name (e.g. `handle')
but should have some similarity in their shapes that allows motions to be transferred
between completely different objects. 


\ian{Going beyond manipulation based on simple semantic classes is a significant challenge which
we address in this work. Instead, we must also make use of visual information (point-clouds), and a 
natural language instruction telling the robot what to do since many possible affordances can exist for the same part.
Designing useful features for either of these modalities alone is already 
difficult, and designing features which combine the two for manipulation purposes is extremely
challenging.}

Obtaining a good common representation between different modalities is
challenging for two main reasons. First, each modality might intrinsically have
very different statistical properties -- for example, most trajectory
representations are inherently dense, while a bag-of-words representation of
language is by nature sparse. This makes it challenging to apply algorithms
designed for unimodal data, as one modality might overpower the others.
Second, even with expert knowledge, it is extremely challenging to design
joint features between such disparate modalities. 
Humans are able to map similar concepts from different sensory system to the same concept using 
\emph{common representation} between different modalities \citep{erdogan2014transfer}.
For example, we are able to correlate the appearance with feel of a banana, 
or a language instruction with a real-world action.
This ability to fuse information from different input modalities
and map them to actions is extremely useful to a household robot.

In this work, we use deep neural networks to learn a shared embedding 
between the combination of object parts in the environment (point-cloud) and natural language instructions, 
and manipulation trajectories (Fig.~\ref{fig:main_fig}). 
This means that all three modalities are projected to the
\emph{same} feature space.
We introduce an algorithm that learns to pull semantically
similar environment/language pairs and their corresponding trajectories 
to the same
regions, and push environment/language pairs away from irrelevant trajectories
based on how irrelevant these trajectories are.
Our algorithm allows for efficient inference because,
given a new instruction and point-cloud,
we only need to 
find the nearest trajectory to the projection of this pair in the learned embedding space, which
can be done using fast 
nearest-neighbor algorithms \citep{flann_pami_2014}.

In the past, deep learning methods have shown impressive results
for learning features in a wide variety of domains 
\citep{krizhevsky2012imagenet,socher2011semi,hadsell2008deep}
and even learning cross-domain embeddings for, for example, language
and image features \citep{srivastava2012multimodal}. 
In contrast to these existing methods, here we present a new
pre-training algorithm for initializing networks to be used for
joint embedding of different modalities. 

Such deep learning algorithms require a large dataset for training. 
However, collecting a large enough dataset 
of expert demonstrations on a large number of objects
is very expensive as 
it requires joint physical presence of the robot, an expert, and the object to be manipulated.
In this work, we show that we can crowd-source the collection of manipulation demonstrations to the
public over the web through our Robobarista platform. 
Since crowd-sourced demonstrations might be noisy and
sub-optimal, we present a new learning algorithm
which handles this noise.
\jae{With our noise handling algorithm, our model trained with crowd-sourced demonstrations
outperforms the model trained with expert demonstrations,
even with the significant amount of noise in crowd-sourced manipulation demonstrations.}

\jae{
Furthermore, in contrast to previous approaches
based on learning from demonstration (LfD) that learn a mapping from a state to
 an action \citep{argall2009survey},
our work complements LfD as we focus on the entire manipulation motion,
as opposed to a sequential state-action mapping.
}


Our Robobarista web-based crowdsourcing platform ({\small \url{http://robobarista.cs.cornell.edu}}) allowed us to
collect a large dataset of \emph{116 objects} 
with \emph{250 natural language instructions}
for which there are \emph{1225 crowd-sourced manipulation trajectories} from 71 non-expert users,
which we use to validate our methods.
We also present experiments on our robot using our approach.
In summary, the key contributions of this work are:
\jae{
\begin{itemize}
\item We present a novel approach to manipulation planning via \textit{part-based transfer} between
different objects that allows manipulation of novel objects.
\item We introduce a new \textit{deep learning model} that handles three modalities with noisy labels from crowd-sourcing.
\item We introduce a new algorithm, which learns an \textit{embedding space} while enforcing
a varying and \textit{loss-based margin}, along with a new unsupervised pre-training method which outperforms standard pre-training algorithms \citep{SAE}.
\item We develop an online platform which allows the  
incorporation of \textit{crowd-sourcing} to manipulation planning and introduces a large-scale manipulation dataset.
\item We evaluate our algorithms on this dataset, showing significant improvement over other state-of-the-art methods.
\end{itemize}
}

%
%



\section{Related Work}

Improving robotic perception and teaching
manipulation strategies to robots has been
a major research area in recent years.
In this section, we describe related work in
various aspects of learning to manipulate novel
objects.

\header{Scene Understanding.}
In recent years, there has been significant research focus on semantic scene understanding
\cite{li2009towards,koppula2011semantic,krizhevsky2012imagenet,wu2014_hierarchicalrgbd},
human activity detection \cite{sung_rgbdactivity_2012,Hu2014humanact},
and features for RGB-D images and point-clouds \cite{socher2012convolutional,lai_icra14}.
Similar to our idea of using part-based transfers, 
the deformable part model \cite{girshick2011object} was effective
in object detection.
However, classification of objects, object parts, or human activities alone
does not provide enough information for a robot to reliably plan manipulation.
Even a simple category such as `kitchen sinks' has a huge amount of variation in how 
different instances are manipulated -- for example, handles and knobs must be manipulated differently,
and different orientations and positions of these parts require very different strategies such as
pulling the handle upwards, pushing upwards, pushing sideways, and so on.
On the other hand, direct perception approaches \cite{gibson1986ecological,kroemer2012kernel}
skip the intermediate object labels
and directly perceive affordances based on the shape of the object.
These works focus on detecting the part known to afford certain actions,
such as `pour,' given the object,
while we focus on predicting the correct motion given the object part.

\header{Manipulation Strategy.}
Most works in robotic manipulation focus on task-specific
manipulation of \emph{known} objects---for example, 
baking cookies with known tools \cite{bollini2011bakebot}
and folding the laundry \cite{miller2012geometric} --
or focus on learning specific motions such as grasping \cite{kehoe2013cloud}
and opening doors \cite{endres2013learning}.
Others \cite{sung_synthesizingsequences_2014,misra2014tellme} focus on sequencing manipulation tasks assuming 
perfect manipulation primitives such as \textit{grasp} and \textit{pour} are available.
Instead, here, we use learning to generalize to manipulating
\emph{novel} objects never seen before by the robot, without relying on preprogrammed 
motion primitives.


For the more general task of manipulating new instances of objects, previous approaches rely 
on finding articulation \cite{sturm2011probabilistic,pillai2014articulated} 
or using interaction \cite{katz2013interactive}, but they are limited by tracking performance of
a vision algorithm.
Many objects that humans operate daily have small parts such as `knobs', which leads to
significant occlusion as manipulation is demonstrated.
Another approach using part-based transfer between objects has been shown to be successful
for grasping \cite{dang2012semantic,detry2013learning}.
We extend this approach and introduce a deep learning model that
enables part-based transfer of \textit{trajectories}
by automatically learning relevant features.
Our focus is on the generalization of manipulation trajectory via part-based transfer
using point-clouds without knowing objects a priori
and without assuming any of the sub-steps (`approach', `grasping', and `manipulation').

\jae{
A few recent works use deep learning approaches for 
robotic manipulation. \citet{levine2015manipulation} use a Gaussian mixture
model to learn system dynamics, then use these to learn 
a manipulation policy using a deep network. 
\citet{lenz2015deepmpc} use a deep network to learn system dynamics for real-time model-predictive control.
Both these works
focus on learning low-level controllers, whereas here we learn
high-level manipulation trajectories.
}



\header{Learning from Demonstration.}
Several successful approaches for teaching robots tasks,
such as helicopter maneuvers \cite{abbeel2010autonomous}
or table tennis \cite{mulling2013learning},
have been based on Learning from Demonstration (LfD) \cite{argall2009survey}.
Although LfD allows end users to demonstrate a manipulation task by simply taking 
control of the robot's arms,
it focuses on learning individual actions and separately relies on high level task composition 
\cite{mangin2011unsupervised, daniel2012learning}
or is often limited to previously seen objects \cite{phillipslearning, pastor2009learning}.
We believe that learning a single model for an action like `turning on'
is impossible because human environments have so many variations.

Unlike learning a model from demonstration,
instance-based learning \cite{aha1991instance,forbes2014robot} replicates
one of the demonstrations.
Similarly, we directly transfer one of the demonstrations, but
focus on generalizing manipulation planning to completely new objects,
enabling robots to manipulate objects they have never seen before.

\header{Metric Embedding.}
Several works in machine learning make use of the power of shared embedding spaces.
LMNN \cite{weinberger2005distance} learns a max-margin Mahalanobis distance 
for a unimodal input feature space.
\citet{weston2011wsabie} learn linear mappings from image and language features to a common embedding 
space for automatic image annotation. \citet{moore2012playlist} learn to map songs and natural language
tags to a shared embedding space.
However, these approaches learn only a shallow, linear mapping from input features, whereas here we learn
a deep non-linear mapping which is less sensitive to input representations.

\header{Deep Learning.}
In recent years, deep learning algorithms have enjoyed huge successes, particularly in the 
domains of
of vision and natural language processing (e.g. \cite{krizhevsky2012imagenet,socher2011semi}).
In robotics, deep learning has previously  been  successfully used
for detecting grasps for novel objects in multi-channel RGB-D images \cite{lenz2013deep}
and for classifying terrain from long-range vision \cite{hadsell2008deep}.


\citet{ngiam2011multimodal} use deep learning to learn features incorporating both video and audio modalities.
\citet{sohn2014improved} propose a new generative learning
algorithm for multimodal data which improves robustness to missing modalities at inference time.
In these works,
a single network takes all modalities as
inputs,
whereas here we perform joint embedding of multiple modalities using multiple networks.

Several previous works use deep networks for joint embedding between different feature spaces.
\citet{mikolov2013translation} map different languages to a joint feature space for translation. 
\citet{srivastava2012multimodal} map images and natural language ``tags'' to the same space for
automatic annotation and retrieval. 
While these works use conventional pre-training algorithms, here we present a new pre-training
approach for learning embedding spaces and show that it outperforms these existing methods (Sec.~\ref{sec:results}.)
Our algorithm trains each layer to map similar
cases to similar areas of its feature space, as opposed to other
methods which either perform variational learning \citep{hinton2006reducing}
or train for reconstruction \citep{SAE}.

%

\citet{hu2014discriminative} also use a deep network for
metric learning for the task of face verification.
Similar to LMNN \cite{weinberger2005distance}, \citet{hu2014discriminative} 
enforces 
a constant margin between
distances among inter-class objects and among intra-class objects.
In Sec.~\ref{sec:results}, we show that our approach, which uses a
loss-dependent variable margin, produces better results for our problem.
\jae{
Our work builds on deep neural network to embed three different modalities of 
point-cloud, language, and trajectory into shared embedding space
while handling lots of label-noise originating from crowd-sourcing.
}


\header{Crowd-sourcing.}
Many approaches to teaching robots manipulation and other skills have relied on demonstrations
by skilled experts
\cite{argall2009survey,abbeel2010autonomous}.
Among previous efforts to scale teaching to the crowd \cite{crick2011human,tellex2014asking,jainsaxena2015_learningpreferencesmanipulation},
\citet{forbes2014robot} employs a similar approach towards crowd-sourcing
but collects
multiple instances of similar table-top manipulation with same object.
Others also build web-based platform for crowd-sourcing manipulation
\cite{toris2013robotsfor,toris2014robot}.
However, these approaches either depend on the presence of an expert (due to required 
special software),
or require a real robot at a remote location.
Our Robobarista platform borrows some components of work from \citet{alexander2012robot},
but works on any standard web browser with OpenGL support and incorporates real point-clouds of various scenes.

\section{Our Approach}

\jae{
Our goal is to build an algorithm that allows a robot to infer a manipulation trajectory
when it is introduced to a new object or appliance and its natural language instruction manual.
The intuition for our approach is that
many differently-shaped objects share similarly-operated object parts; 
thus, the manipulation trajectory of an object can be transferred to a
completely different object if they share similarly-operated parts.
}



\jae{
For example, the motion required to operate the handle of the espresso machine 
in Figure~\ref{fig:espresso_transfer} is almost identical to the motion required
to flush a urinal with a handle.
By identifying and transferring trajectories from prior experience with parts of other objects,
robots can manipulate even objects they have never seen before.
}

\jae{
We first formulate this problem as a structured prediction problem
as shown in Figure~\ref{fig:espresso_transfer}.
Given a point-cloud for each part of an espresso machine and
a natural language instruction such as `Push down on the handle to add hot water',
our algorithm outputs a trajectory which executes the task, 
using a pool of prior motion experience.
}

\jae{
This is a challenging problem because the object is entirely new to the
robot, and because it must jointly consider the point-cloud, natural
language instruction, and each potential trajectory.
Moreover, manually designing useful features from these three modalities 
is extremely challenging.
}

\jae{
We introduce a \textit{deep multimodal embedding} approach 
that learns a shared, semantically meaningful embedding space
between these modalities,
while dealing with a noise in crowd-sourced demonstration data.
Then, we introduce our Robobarista crowd-sourcing platform, which allows us
to easily scale the collection of manipulation demonstrations to 
non-experts on the web.
}

\subsection{Problem Formulation}
\label{sec:prob_form}

Our goal is to learn a function $f$ that maps a given pair of point-cloud 
$p \in \mathcal{P}$ of an object part
and a natural language instruction $l \in \mathcal{L}$ to a trajectory $\tau \in 
\mathcal{T}$ that can manipulate the object part as described by free-form natural language $l$:
\begin{equation}
f : \mathcal{P} \times \mathcal{L} \rightarrow \mathcal{T}  \label{eqn:f}
\end{equation}
\jae{For instance, given the handle of the espresso machine in Figure~\ref{fig:espresso_transfer}
and an natural language instruction `Push down on the handle to add hot water',
the algorithm should output a manipulation trajectory that will correctly accomplish
the task on the object part according to the instruction.
}

\noindent
\textbf{Point-cloud Representation.}
Each instance of a point-cloud $p \in \mathcal{P}$ is represented as a set of $n$ points in three-dimensional
Euclidean space where each point $(x,y,z)$ is represented with its RGB color $(r,g,b)$:
$$p = \{p^{(i)} \}^n_{i=1} = {\{(x,y,z,r,g,b)^{(i)}\}}^n_{i=1}$$
The size of this set varies for each instance.
These points are often obtained by stitching together a sequence of sensor data 
from an RGBD sensor \citep{izadi2011kinectfusion}.

\noindent
\textbf{Trajectory Representation.}
\label{sec:trajrep}
Each trajectory $\tau \in \mathcal{T}$ is represented as a sequence of $m$ \textit{waypoints},
where each waypoint consists of gripper status $g$, translation $(t_x, t_y, t_z)$,
and rotation $(r_x, r_y, r_z, r_w)$ with respect to the origin:
$$\tau= \{\tau^{(i)}\}^m_{i=1} = {\{(g, t_x, t_y, t_z, r_x, r_y, r_z, r_w)^{(i)}\}}^m_{i=1}$$
where  $g \in \{\text{``open''},\text{``closed''},\text{``holding''}\}$.
$g$ depends on the type of the end-effector,
which we have assumed to be a two-fingered parallel-plate gripper like that of PR2 or Baxter.
The rotation is represented as quaternions $(r_x, r_y, r_z, r_w)$
instead of the more compact Euler angles to prevent problems such as
gimbal lock. 

\noindent
\textbf{Smooth Trajectory.} To acquire a smooth trajectory from
a waypoint-based trajectory $\tau$, 
we interpolate intermediate waypoints.
Translation is linearly interpolated and the quaternion is interpolated
using spherical linear interpolation (Slerp) \citep{shoemake1985animating}.


\subsection{Data Pre-processing}
\label{sec:preprocessing}


Each of the point-cloud, language, and trajectory $(p, l, \tau)$ can have any length. 
Thus, we fit raw data from each modality into a fixed-length vector.

We represent point-cloud $p$ of any arbitrary length
as an occupancy grid where each cell 
indicates whether any point lives in the space it represents.
Because point-cloud $p$ consists of only the part of an object which is limited in size,
we can represent $p$ using two occupancy grid-like structures of size $10 \times 10 \times 10$ voxels with different scales: 
one with each voxel spanning a cube of $1 \times 1 \times 1 (cm)$ and the other with 
each cell representing $2.5\times 2.5 \times 2.5 (cm)$.

Each language instruction is represented as a fixed-size bag-of-words representation with
stop words removed.
Finally, for each trajectory $\tau \in \mathcal{T}$, 
we first compute its smooth interpolated trajectory $\tau_s \in \mathcal{T}_s$
(Sec.~\ref{sec:trajrep}),
and then normalize all trajectories $\mathcal{T}_s$ to the same length
while preserving the sequence of gripper states such as `opening', `closing', and `holding'.



\subsection{Direct manipulation trajectory transfer}
\label{sec:transfer_adapt}

Even if we have a trajectory to transfer,
a conceptually transferable trajectory is not necessarily directly compatible
if it is represented with respect to an inconsistent reference point.

To make a trajectory compatible with a new situation without modifying the 
trajectory,
we need a representation method for trajectories, based on point-cloud information,
that allows a \textit{direct
transfer of a trajectory without any modification}.

\noindent
\textbf{Challenges.}
Making a trajectory compatible when transferred to a different object or to a
different instance of the same object without modification
 can be challenging depending on the representation
of trajectories and the variations in the location of the object, 
given in point-clouds. 


\ian{
Many approaches which control high degree of freedom arms such as those of PR2 or Baxter
use configuration-space trajectories, which store a time-parameterized series of joint angles
\citep{thrun2005probabilistic}. While such approaches allow for direct control of joint angles during
control, they require costly recomputation for even a small change in an object's position or orientation.
}

One approach that allows execution without modification 
is representing trajectories with respect to the object
by aligning via point-cloud registration (e.g. \citep{forbes2014robot}). 
However, a large object such as a stove might have many parts (e.g. knobs and handles) whose
positions might vary between different stoves. Thus, object-aligned manipulation of these
parts would not be robust to different stoves, and in general would impede transfer between different
instances of the same object.

Lastly, it is even more challenging if two objects require similar trajectories, but have slightly different shapes.
And this is made more difficult by limitations of the point-cloud data. 
As shown in left of Fig.~\ref{fig:espresso_transfer},
the point-cloud data, even when stitched from multiple angles, are very noisy 
compared to the RGB images.

\noindent
\textbf{Our Solution.}
Transferred trajectories become compatible across different objects when trajectories are 
represented
1) in the task space rather than the configuration space, and 
2) relative to the object \emph{part} in question (aligned based on its principal axis), rather than the object as a whole.

Trajectories can be represented in the task space by recording only the position and orientation
of the end-effector.
By doing so, we can focus on the actual interaction between the robot and the 
environment
rather than the movement of the arm.
It is very rare that the arm configuration affects the completion of the task as long as there is no collision.
With the trajectory represented as a sequence of gripper position and orientation, 
the robot can find its arm configuration that 
is collision free with the environment using inverse kinematics.

However, representing the trajectory in task space is not enough to make transfers
compatible.
The trajectory must also be represented in a common coordinate frame regardless of the object's orientation and shape. 

Thus, we align the negative $z$-axis along gravity and align the $x$-axis along the
principal axis of the object \textit{part} using PCA \citep{hsiao2010contact}.
With this representation, even when the object part's position and orientation changes,
the trajectory does not need to change.
The underlying assumption is that similarly operated object parts share 
similar shapes leading to a similar direction in their principal axes.



\begin{figure}
  \begin{center}
    \includegraphics[width=0.99\columnwidth]{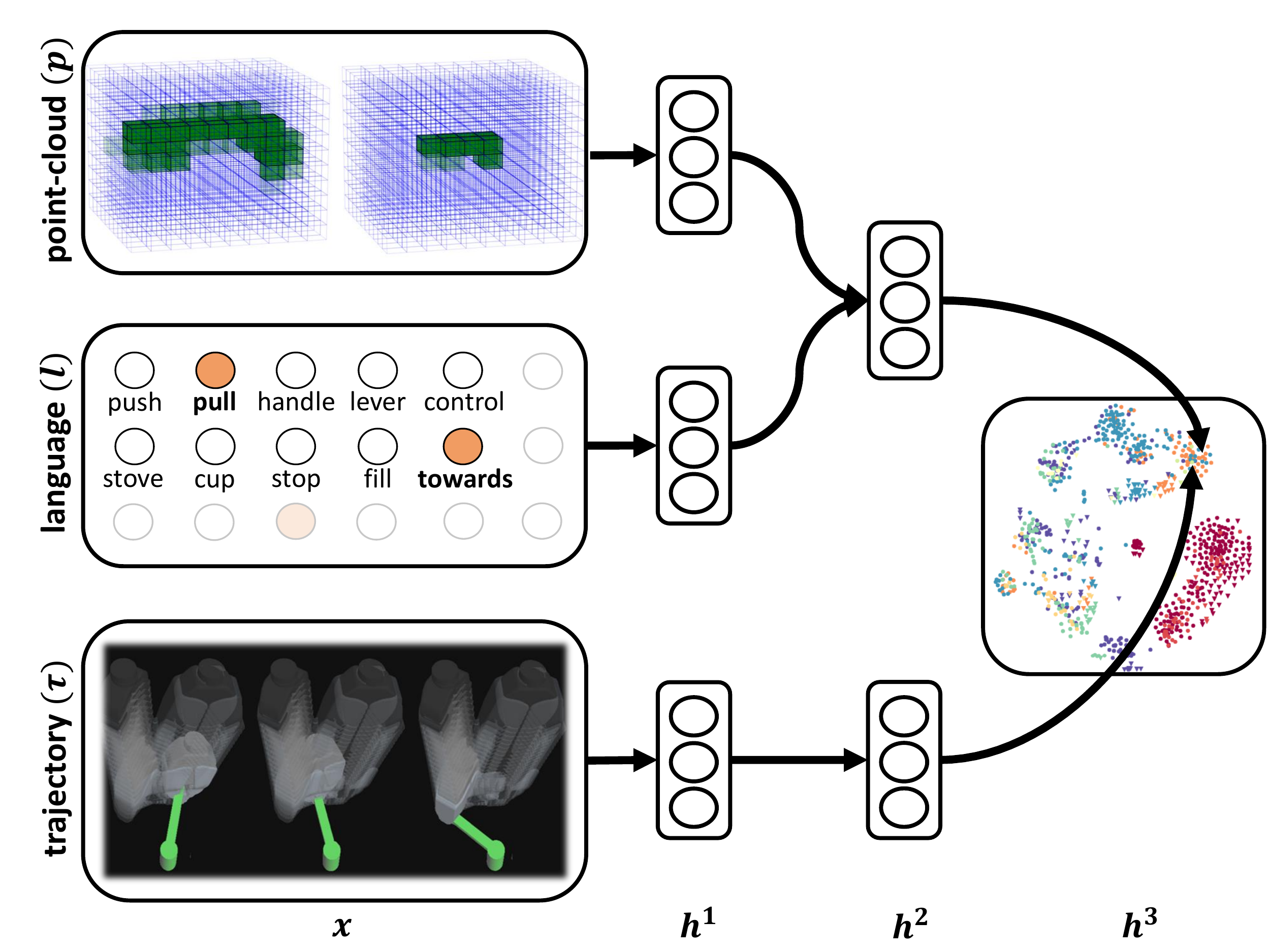}
  \end{center}
  \caption{
    \textbf{Deep Multimodal Embedding:} Our deep neural network learns 
    to embed both point-cloud/natural language instruction combinations and
    manipulation trajectories in the same space. 
    This allows for fast selection of a new trajectory by projecting
    a new environment/instruction pair and choosing its nearest-neighbor trajectory in this space.
    }
  \label{fig:main_fig}
\end{figure}

\section{Deep Multimodal Embedding}
\label{sec:deep_embedding}



\ian{
In this work, we use deep learning to find the most appropriate trajectory
for a given point-cloud and natural language instruction. This is much more challenging
than the uni-modal binary or multi-class classification/regression problems 
(e.g. image recognition \cite{krizhevsky2012imagenet}) to which deep learning has mostly been
applied \citep{bengio2013representation}. We could simply convert our problem 
into a binary classification problem, i.e. ``does this trajectory match this point-cloud/language pair?'' Then, we can use a multimodal feed-forward deep network
\cite{ngiam2011multimodal} to solve this problem \cite{sung_robobarista_2015}.
}

\ian{
However, this approach has several drawbacks. First, it requires evaluating the network
over \emph{every} combination of potential candidate manipulation trajectory and the
given point-cloud/language pair, which is computationally expensive. Second, this 
method does nothing to handle noisy labels, a significant problem when dealing with
crowd-sourced data as we do here. Finally, while this method is capable of producing
reasonable results for our problem, we show in Section~\ref{sec:results} that
a more principled approach to our problem is able to improve over it.
}

\ian{
Instead, in this work, we present a new deep architecture and algorithm which is a
better fit for this problem, directly addressing the challenges inherent in multimodal
data and the noisy labels obtained from crowdsourcing.
}
We learn a joint embedding of point-cloud, language, and trajectory data
into the same low dimensional space.   
We learn non-linear embeddings using a deep learning approach
which maps raw data from these three different modalities to a joint embedding space.

We then use this space to find known
trajectories which are good matches for new
combinations of object parts and instructions.
Compared to previous work  
that exhaustively runs a full network over all these combinations \citep{sung_robobarista_2015}, 
our approach allows us to pre-embed all candidate trajectories into this common feature space.
Thus, the most appropriate trajectory can be identified 
by embedding only a new point-cloud/language pair and then finding its nearest neighbor.

In our joint feature space, proximity between two mapped points 
should reflect how relevant two data-points are to each other,
even if they are from completely different modalities.
We train our network to bring demonstrations that would manipulate a given object 
according to some language instruction closer to the mapped point for that object/instruction pair,
and to push away demonstrations that would not correctly manipulate the object.
Trajectories which have no semantic relevance to the object are pushed much further 
away than trajectories that have some relevance, 
even if the latter would not manipulate the object according to the instruction.

Prior to learning a full joint embedding of all three modalities,
we pre-train embeddings of subsets of the modalities 
to learn semantically meaningful embeddings for these modalities,
leading to improved performance as shown in Section ~\ref{sec:results}.



\jae{
To solve this problem of learning to manipulate novel objects and appliance
as defined in equation~(\ref{eqn:f}),
}
we learn two different mapping functions
that map to a common space---one from a point-cloud/language pair 
and the other from a trajectory.
More formally, we want to learn $\Phi_{\mathcal{P},\mathcal{L}}(p,l)$ and
$\Phi_{\mathcal{T}}(\tau)$ which map to a joint feature space $\mathbb{R}^M$:
\begin{align*}
\Phi_{\mathcal{P},\mathcal{L}}(p,l)&: (\mathcal{P},\mathcal{L}) \rightarrow \mathbb{R}^M  \\
\Phi_{\mathcal{T}}(\tau)&: \mathcal{T} \rightarrow \mathbb{R}^{M}
\end{align*}
Here, we represent these mappings with a deep neural network, 
as shown in Figure~\ref{fig:main_fig}.

The first, $\Phi_{\mathcal{P},\mathcal{L}}$, which maps point-clouds and trajectories, 
is defined as a combination of two mappings. The first of these maps to a joint 
point-cloud/language space $\mathbb{R}^{N_{2,pl}}$ ---
$\Phi_{\mathcal{P}}(p):\mathcal{P} \rightarrow \mathbb{R}^{N_{2,pl}}$ and
$\Phi_{\mathcal{L}}(l):\mathcal{L} \rightarrow \mathbb{R}^{N_{2,pl}}$.
Once each is mapped to $\mathbb{R}^{N_{2,pl}}$, 
this space is then mapped to the joint space shared with trajectory information: 
$\Phi_{\mathcal{P},\mathcal{L}}(p,l): ((\mathcal{P}, \mathcal{L}) \rightarrow \mathbb{R}^{N_{2,pl}}) \rightarrow \mathbb{R}^M$.

\subsection{Model}


We use two separate multi-layer deep neural networks,
one for 
$\Phi_{\mathcal{P},\mathcal{L}}(p,l)$
and one for $\Phi_{\mathcal{T}}(\tau)$.
Take $N_p$ as the size of point-cloud input $p$,
$N_l$ as similar for natural language input $l$, $N_{1,p}$ and
$N_{1,l}$ as the number of hidden units in the first hidden
layers projected from point-cloud and natural language features,
respectively, and $N_{2,pl}$ as the number of hidden units
in the combined point-cloud/language layer. With $W$'s as network
weights, which are the learned parameters of our system, and
$a(\cdot)$ as a rectified linear unit (ReLU)
activation function \citep{zeiler2013rectified},
our model for projecting from point-cloud and language features
to the shared embedding $h^3$ is as follows:
\begin{align*}
h^{1,p}_i & = a\left(\textstyle \sum_{j=0}^{N_p} W^{1,p}_{i,j}  p_j \right)\\
h^{1,l}_i & = a\left(\textstyle \sum_{j=0}^{N_l} W^{1,l}_{i,j}  l_j \right)\\
h^{2,pl}_i & = a\left(\textstyle \sum_{j=0}^{N_{1,p}} W^{2,p}_{i,j} h^{1,p}_j + \sum_{j=0}^{N_{1,l}} W^{2,l}_{i,j} h^{1,l}_j\right) \\
h^{3}_i & = a\left(\textstyle \sum_{j=0}^{N_{2,pl}} W^{3,pl}_{i,j} h^{2,pl}_j \right)
\end{align*}
The model for projecting from trajectory input $\tau$ is similar,
except it takes input only from a single modality.

\subsection{Inference.}
Once all mappings are learned, we solve the original problem from equation~(\ref{eqn:f})
by choosing, from a library of prior trajectories, the trajectory 
that gives the highest similarity (closest in distance)
to the given point-cloud $p$ and language $l$ in our joint embedding space $\mathbb{R}^M$.
As in previous work \citep{weston2011wsabie}, similarity is defined as $sim(a,b) = a \cdot b$, and
the trajectory that maximizes the magnitude of similarity is selected:
$$\argmax_{\tau \in \mathcal{T}} sim(\Phi_{\mathcal{P},\mathcal{L}}(p,l), \Phi_{\mathcal{T}}(\tau))$$

The previous approach to this problem \citep{sung_robobarista_2015} 
required projecting the combination of the current point-cloud and 
natural language instruction with \emph{every} trajectory in the training set through the network during inference.
Here, we pre-compute the representations of all training trajectories
in $h^3$, and need only project the new point-cloud/language pair to $h^3$ and find its nearest-neighbor
trajectory in this embedding space. As shown in 
Section~\ref{sec:results}, this significantly improves 
both the runtime and accuracy of
our approach and makes it much more scalable to larger training datasets like those collected with 
crowdsourcing platforms. 

\section{Learning Joint Point-cloud/Language/Trajectory Model}

The main challenge of our work is to learn a model which maps three disparate
modalities -- point-clouds, natural language, and trajectories -- to a single
semantically meaningful space.
We introduce a method that learns a common point-cloud/language/trajectory space such that 
all trajectories relevant to a given task (point-cloud/language combination) 
should have higher similarity to the projection of that task than 
task-irrelevant trajectories.
Among these irrelevant trajectories, some might be less relevant than others,
and thus should be pushed further away.

For example, given a door knob that needs to be grasped normal to the door surface and 
an instruction to rotate it clockwise, 
a trajectory that correctly approaches the door knob but rotates counter-clockwise
should have higher similarity to the task than one
which approaches the knob from a completely incorrect angle and does not execute any rotation.

\jae{
For every training point-cloud/language pair $(p_i, l_i)$, 
we have two sets of demonstrations:
a set of trajectories $\mathcal{T}_{i,S}$ that are relevant (similar) 
to this task and a set of trajectories $\mathcal{T}_{i,D}$ that are irrelevant (dissimilar)
as described in Sec.~\ref{sec:noise}.
}
For each pair of $(p_i, l_i)$,
we want all projections of $\tau_j \in \mathcal{T}_{i,S}$ to have higher similarity
to the projection of $(p_i, l_i)$ than $\tau_k \in \mathcal{T}_{i,D}$.
A simple approach would be to train the network
to distinguish these two sets by enforcing
a finite distance (safety margin) between the similarities of
these two sets \citep{weinberger2005distance},
which can be written in the form of a constraint:
$$sim(\Phi_{\mathcal{P},\mathcal{L}}(p_i,l_i), \Phi_{\mathcal{T}}(\tau_j)) 
\geq 1 + sim(\Phi_{\mathcal{P},\mathcal{L}}(p_i,l_i), \Phi_{\mathcal{T}}(\tau_k))$$

Rather than simply being able to distinguish two sets, we want to learn
semantically meaningful embedding spaces from different modalities.
Recalling our earlier example 
where one incorrect trajectory for manipulating
a door knob was much closer to correct than another, it is clear that our
learning algorithm should drive some of incorrect trajectories to be more
dissimilar than others.
The difference between the similarities of $\tau_j$ and $\tau_k$ to the projected
point-cloud/language pair $(p_i, l_i)$ should be at least the loss $\Delta(\tau_j,\tau_k)$.
This can be written as a form of a constraint:
\begin{align*}
\forall \tau_j \in \mathcal{T}_{i,S},  \forall \tau_k  \in & \; \mathcal{T}_{i,D} \\
sim(\Phi_{\mathcal{P},\mathcal{L}}(p_i,l_i)&, \Phi_{\mathcal{T}}(\tau_j)) \\
&\geq \Delta(\tau_j, \tau_k) + sim(\Phi_{\mathcal{P},\mathcal{L}}(p_i,l_i), \Phi_{\mathcal{T}}(\tau_k)) \\
\end{align*}
Intuitively, this forces trajectories with higher DTW-MT distance from the ground truth to embed further 
than those with lower distance. Enforcing all combinations of these constraints could grow exponentially large.
Instead, similar to the cutting plane method for structural support vector machines \citep{tsochantaridis2005large},
we find the most violating trajectory $\tau' \in \mathcal{T}_{i,D}$
for each training pair of $(p_i, l_i, \tau_i \in \mathcal{T}_{i,S})$
at each iteration.
The most violating trajectory 
has the highest similarity augmented with the loss scaled by a constant $\alpha$:
$$\tau'_i = \argmax_{\tau \in \mathcal{T}_{i,D}} (sim(\Phi_{\mathcal{P},\mathcal{L}}(p_i,l_i), \Phi_{\mathcal{T}}(\tau)) 
+ \alpha \Delta(\tau_i, \tau))$$

The cost of our deep embedding space $h^3$ is computed 
as the hinge loss of the most violating trajectory.
\begin{align*}
L_{h^3}(p_i, l_i, \tau_i) = |\Delta(\tau'_i, \tau_i) + & sim(\Phi_{\mathcal{P},\mathcal{L}}(p_i,l_i), \Phi_{\mathcal{T}}(\tau'_i))  \\
- & sim(\Phi_{\mathcal{P},\mathcal{L}}(p_i,l_i), \Phi_{\mathcal{T}}(\tau_i))|_+
\end{align*}

The average cost of each minibatch is back-propagated 
through all the layers of the deep neural network
using the AdaDelta \citep{zeiler2012adadelta} algorithm.

\begin{figure}
  \begin{center}
    \includegraphics[width=\columnwidth]{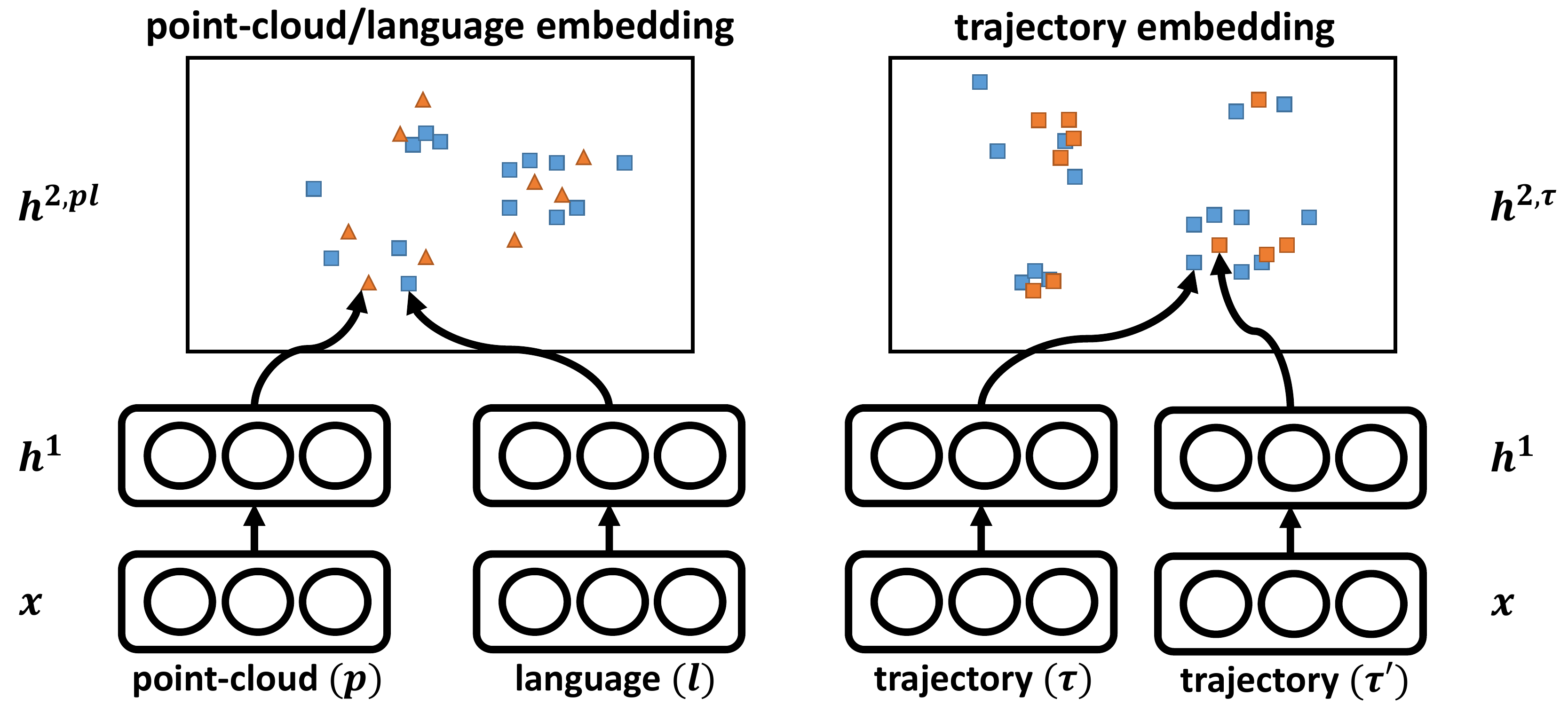}
  \end{center}
  \vskip -.07 in
  \caption{\textbf{Pre-training lower layers:} Visualization of our pre-training approaches for
  $h^{2,pl}$ and $h^{2,\tau}$. For $h^{2,pl}$, our 
  algorithm pushes matching point-clouds and instructions
  to be more similar. For $h^{2,\tau}$, our algorithm
  pushes trajectories with higher DTW-MT similarity to
  be more similar.}
  \label{fig:joint_embedding_pretraining}
\end{figure}

\subsection{Pre-training Joint Point-cloud/Language Model}
\label{sec:pre_task}

One major advantage of modern deep learning methods is the use of unsupervised pre-training
to initialize neural network parameters to a good starting point before the final supervised
fine-tuning stage. 
Pre-training helps these high-dimensional networks
to avoid overfitting to the training data.

Our lower layers $h^{2,pl}$ and $h^{2,\tau}$ represent features extracted
exclusively from the combination of point-clouds and language,  
and from trajectories, respectively.
Our pre-training method initializes $h^{2,pl}$ and $h^{2,\tau}$
as semantically meaningful embedding spaces similar to $h^3$,
as shown later in Section~\ref{sec:results}.

First, we pre-train the layers leading up to these layers using 
spare de-noising autoencoders \citep{vincent2008extracting,zeiler2013rectified}.
Then, our process for pre-training $h^{2,pl}$ is similar to 
our approach to fine-tuning a semantically meaningful embedding space for $h^3$
presented above,
except now we find the most violating language $l'$
while still relying on a loss over the associated optimal trajectory:
$$l' = \argmax_{l \in \mathcal{L}} (sim(\Phi_{\mathcal{P}}(p_i), \Phi_{\mathcal{L}}(l)) + \alpha \Delta(\tau, \tau_i^*))$$
\begin{align*}
L_{h^{2,pl}}(p_i, l_i, \tau_i) = |\Delta(\tau_i, \tau') + & sim(\Phi_{\mathcal{P}}(p_i), \Phi_{\mathcal{L}}(l'))  \\
- & sim(\Phi_{\mathcal{P}}(p_i), \Phi_{\mathcal{L}}(l_i)) |_+
\end{align*}
Notice that although we are training this embedding space to project from point-cloud/language data, 
we guide learning using trajectory information.


After the projections $\Phi_{\mathcal{P}}$ and $\Phi_{\mathcal{L}}$ are tuned,
the output of these two projections are added to form the output of layer $h^{2,pl}$ in the final feed-forward network.



\subsection{Pre-training Trajectory Model}
\label{sec:pre_traj}

For our task of inferring manipulation trajectories for novel objects, it is especially important
that similar trajectories $\tau$ map to similar regions in the feature space defined by $h^{2,\tau}$, 
so that trajectory embedding $h^{2,\tau}$ itself is semantically meaningful
and they can in turn be mapped to similar regions in $h^3$. 
Standard pretraining methods, such as sparse de-noising autoencoder \citep{vincent2008extracting,zeiler2013rectified}
would only pre-train $h^{2,\tau}$ to reconstruct individual trajectories. 
Instead, we employ pre-training similar to Sec.~\ref{sec:pre_task}, 
except now we pre-train for only a single modality -- trajectory data.

As shown on right hand side of Fig.~\ref{fig:joint_embedding_pretraining},
the layer that embeds to $h^{2,\tau}$ is duplicated.
These duplicated embedding layers are treated as if they were two different
modalities,
but all their weights are shared and updated simultaneously. 
For every trajectory $\tau \in \mathcal{T}_{i,S}$, 
we can again find the most violating $\tau' \in \mathcal{T}_{i,D}$ 
and the minimize a similar cost function as we do for $h^{2,pl}$.

\subsection{Label Noise}
\label{sec:noise}


When our data contains a significant number of noisy trajectories $\tau$, e.g. due to crowd-sourcing (Sec.~\ref{sec:crowdsourcing}), not all
trajectories should be trusted as equally appropriate, as will be shown
in Sec.~\ref{sec:experiments}.

For every pair of inputs $(p_i,l_i)$, we have 
$\mathcal{T}_i=\{\tau_{i,1}, \tau_{i,2}, ..., \tau_{i,n_i} \}$, 
a set of trajectories submitted by the crowd for $(p_i,l_i)$. 
First, the best candidate label $\tau^*_i \in \mathcal{T}_i$ for $(p_i,l_i)$ is selected as the one with the
smallest average trajectory distance to the others:
$$\tau^*_i = \argmin_{\tau \in \mathcal{T}_i} \frac{1}{n_i} \sum_{j=1}^{n_i} \Delta(\tau, \tau_{i,j}) $$
We assume that at least half of the crowd tried to give a reasonable demonstration.
Thus a demonstration with the smallest average distance to all other demonstrations
must be a good demonstration.
\jae{
We use the DTW-MT distance function (described later in Sec.~\ref{sec:metric})
for our loss function $\Delta(\tau, \bar{\tau})$,
but it could be replaced by any function that computes the loss of predicting $\bar{\tau}$
when $\tau$ is the correct demonstration.
}

\jae{
Using the optimal demonstration and a loss function $\Delta(\tau, \bar{\tau})$
for comparing demonstrations, 
we find a set of trajectories $\mathcal{T}_{i,S}$ that are relevant (similar)
to this task and a set of trajectories $\mathcal{T}_{i,D}$ that are irrelevant (dissimilar.)
We can use thresholds $(t_S, t_D)$ determined by the expert
to generate two sets from the pool of trajectories:
$$\mathcal{T}_{i,S} = \{\tau \in \mathcal{T} | \Delta(\tau_i^*, \tau) < t_S \}$$
$$\mathcal{T}_{i,D} = \{\tau \in \mathcal{T} | \Delta(\tau_i^*, \tau) > t_D \}$$
}

\jae{
This method allows our model to be robust against noisy labels
and also serves as a method of data augmentation 
by also considering demonstrations given for other tasks 
in both sets of $\mathcal{T}_{i,S}$ and  $\mathcal{T}_{i,D}$.
}

\begin{figure*}[tb]
\begin{center}
\includegraphics[width=\textwidth]{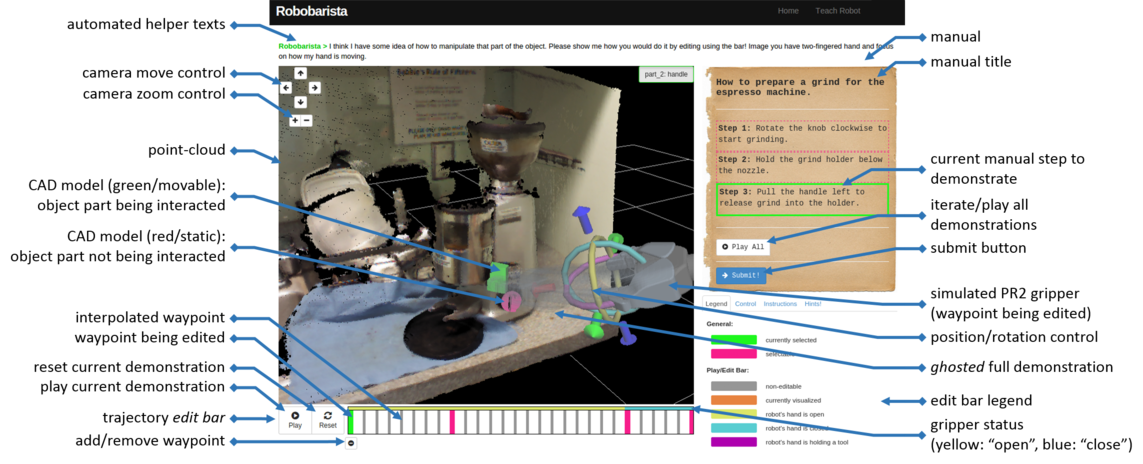}
\end{center}
\vskip -0.2in
\caption{
\textbf{Screen-shot of Robobarista,} the crowd-sourcing platform running on Chrome browser.
We have built Robobarista platform for collecting a large number of crowd demonstrations for teaching the robot.
}
\label{fig:robobarista}
\end{figure*}

\section{Loss Function for Manipulation Trajectory}
\label{sec:metric}

For both learning and evaluation, we need a function
which accurately represents distance between two
trajectories.
Prior metrics for trajectories consider only their translations
(e.g. \citep{koppula2013_anticipatingactivities})
and not their rotations \emph{and} gripper status.
We propose a new measure, which uses dynamic time warping, for evaluating manipulation trajectories.
This measure non-linearly warps two trajectories of 
arbitrary lengths to produce a matching, then computes
cumulative distance as the sum of cost of all matched waypoints.
The strength of this measure
is that weak ordering is maintained among matched waypoints and that every waypoint 
contributes to the cumulative distance.

For two trajectories of arbitrary lengths,
$\tau_A = \{\tau_A^{(i)}\}_{i=1}^{m_A}$ and
$\tau_B = \{\tau_B^{(i)}\}_{i=1}^{m_B}$
, we define a matrix $D \in \mathbb{R}^{m_A \times m_B}$, 
where $D(i, j)$ is the cumulative distance of an optimally-warped matching
between trajectories up to index $i$ and $j$, respectively, of each trajectory.
The first column and the first row of
$D$ is initialized as:
\begin{align*}
D(i,1) &= \sum_{k=1}^{i}c(\tau_A^{(k)}, \tau_B^{(1)})\;\;\; \forall i \in [1, m_A] \\
D(1,j) &= \sum_{k=1}^{j}c(\tau_A^{(1)}, \tau_B^{(k)})\;\;\; \forall j \in [1, m_B] 
\end{align*}
where $c$ is a local cost function between two waypoints (discussed later). The rest of $D$
is completed using dynamic programming:
\begin{align*}
D(i,j) = & c(\tau_A^{(i)}, \tau_B^{(j)}) \\
         & +  \min\{D(i-1, j-1), D(i-1,j), D(i, j-1)\}
\end{align*}


Given the constraint that $\tau_A^{(1)}$ is matched to $\tau_B^{(1)}$,
the formulation ensures that every waypoint contributes to the
final cumulative distance $D(m_A, m_B)$. Also,
given a matched pair $(\tau_A^{(i)}, \tau_B^{(j)})$, no waypoint
preceding $\tau_A^{(i)}$ is matched to a waypoint succeeding $\tau_B^{(j)}$,
encoding weak ordering. 

The pairwise cost function $c$ between matched waypoints $\tau_A^{(i)}$ and $\tau_B^{(j)}$
is defined:

{
\vskip -.1in
\small
\begin{align*}
c(\tau_A^{(i)}, \tau_B^{(j)}; \alpha_T, \alpha_R, & \beta, \gamma) =  w(\tau_A^{(i)};\gamma) w(\tau_B^{(j)};\gamma)  \\
\bigg(\frac{d_T(\tau_A^{(i)}, \tau_B^{(j)})}{\alpha_T}& + \frac{d_R(\tau_A^{(i)}, \tau_B^{(j)})}{\alpha_R}\bigg) \bigg(1 + \beta d_G(\tau_A^{(i)}, \tau_B^{(j)}) \bigg) \\
\text{where} \hspace{5 mm} d_T(\tau_A^{(i)},\tau_B^{(j)}) &= ||(t_x, t_y, t_z)_{A}^{(i)} - (t_x, t_y, t_z)_{B}^{(j)}||_2  \\  
d_R(\tau_A^{(i)},\tau_B^{(j)})  &= \text{angle difference between $\tau_{A}^{(i)}$ and $\tau_B^{(j)}$} \\
\qquad d_G(\tau_A^{(i)},\tau_B^{(j)}) &= \mathds{1}(g_A^{(i)}=g_B^{(j)}) \\
w(\tau^{(i)}; \gamma) &= exp(-\gamma \cdot ||\tau^{(i)}||_2) 
\end{align*}
\vskip -.05in
}
\noindent
The parameters $\alpha, \beta$ are for scaling translation and rotation errors, and
gripper status errors, respectively. $\gamma$ weighs the importance of a waypoint based on
its distance to the object part.
\footnote{\jae{In this work, we assign $\alpha_T, \alpha_R, \beta, \gamma$ values of $0.0075$ meters, $3.75^{\circ}$, $1$ and $4$ respectively.}}
Finally, as trajectories vary in length,
we normalize $D(m_A, m_B)$ by the number of waypoint pairs that contribute
to the cumulative sum, $|D(m_A, m_B)|_{path^*}$ 
(i.e. the length of the optimal warping path), giving the final form:
\begin{align*}
distance(\tau_A, \tau_B) = \frac{D(m_A, m_B)}{|D(m_A, m_B)|_{path^*}}
\end{align*}
This distance function is used for noise-handling in our model
and as the final evaluation metric.

\section{Robobarista: crowd-sourcing platform}
\label{sec:crowdsourcing}

In order to collect a large number of manipulation demonstrations from the crowd, we built a crowd-sourcing web platform that we call Robobarista (see Fig.~\ref{fig:robobarista}).
It provides a virtual environment where non-expert users can teach robots
via a web browser, without expert guidance or  physical presence with a robot and a target object.

The system simulates a situation where the user encounters 
a previously unseen target object and a natural language instruction manual
for its manipulation.
Within the web browser, users are shown a point-cloud in the 3-D viewer on the left
and a \textit{manual} on the right.
A manual may involve several instructions,
such as ``Push down and pull the handle to open the door''.
The user's goal is to demonstrate how to manipulate
the object in the scene for each instruction.

The user starts by selecting one of the instructions on the right to demonstrate (Fig.~\ref{fig:robobarista}).
Once selected, the target object part
is highlighted and the trajectory \textit{edit bar} appears below the 3-D viewer.
Using the \textit{edit bar}, which works like a video editor, the user can 
playback and edit the demonstration.
The trajectory representation,
as a set of waypoints (Sec.~\ref{sec:trajrep}), is directly shown on the \textit{edit bar}.
The bar shows not only the set of waypoints (red/green) but also the interpolated waypoints (gray).
The user can click the `play' button or hover the cursor over the edit bar 
to examine the current demonstration.
The blurred trail of the current trajectory (\textit{ghosted}) demonstration is
also shown in the 3-D viewer to show its full expected path.


Generating a full trajectory from scratch can be difficult for non-experts. 
Thus, similar to  \citet{forbes2014robot}, we provide a trajectory that
the system has already seen for another object as the initial starting trajectory to 
edit.\footnote{We have made sure that it does not initialize with trajectories from other
folds to keep \textit{5-fold cross-validation} in experiment section valid.}

In order to simulate a realistic experience of manipulation,
instead of simply showing a static point-cloud,
we have overlaid CAD models for parts such as `handle'
so that functional parts actually move as the user tries to manipulate the object.

A demonstration can be edited by: 1) modifying the position/orientation
of a waypoint, 2) adding/removing a waypoint,
and 3) opening/closing the gripper.
Once a waypoint is selected, the PR2 gripper is shown with six directional arrows 
and three rings, used to modify the gripper's position and orientation, respectively.
To add extra waypoints,
the user can hover the cursor over an  interpolated  (gray) 
waypoint on the \textit{edit bar} and click the plus(+) button.
To remove an existing waypoint, the user can hover over it on the \textit{edit bar}
and click minus(-) to remove.
As modification occurs, the edit bar and ghosted demonstration are updated with a new interpolation.
Finally, for editing the status (open/close) of the gripper, the user can simply click on the gripper.

For broader accessibility, all functionality of Robobarista, 
including 3-D viewer, is built using Javascript and WebGL.
\jae{We have made the platform available online
({\small \url{http://robobarista.cs.cornell.edu}})
}

\begin{figure*}[tb]
\begin{center}
\includegraphics[width=\textwidth]{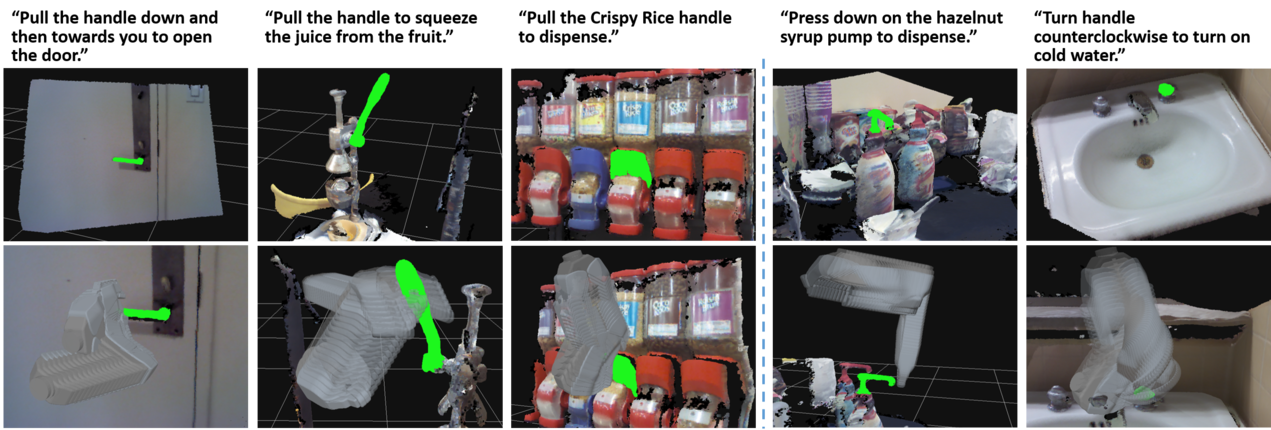}
\end{center}
\caption{
\textbf{Examples from our dataset,} each of which  
consists of a natural language instruction
(top), an object part in point-cloud representation (highlighted), and a manipulation
trajectory (below) collected via Robobarista. 
Objects range from kitchen appliances such as stove and rice cooker to
urinals and sinks in restrooms. As our trajectories are collected from non-experts,
they vary in quality from being likely to complete the manipulation task
successfully (left of dashed line) to being unlikely to do so successfully
(right of dashed line).
}
\label{fig:data_ex}
\end{figure*}


\section{Experiments}
\label{sec:experiments}

\subsection{Robobarista Dataset}
\label{sec:data}

\begin{figure*}[tb]
\begin{center}
  \includegraphics[width=\textwidth]{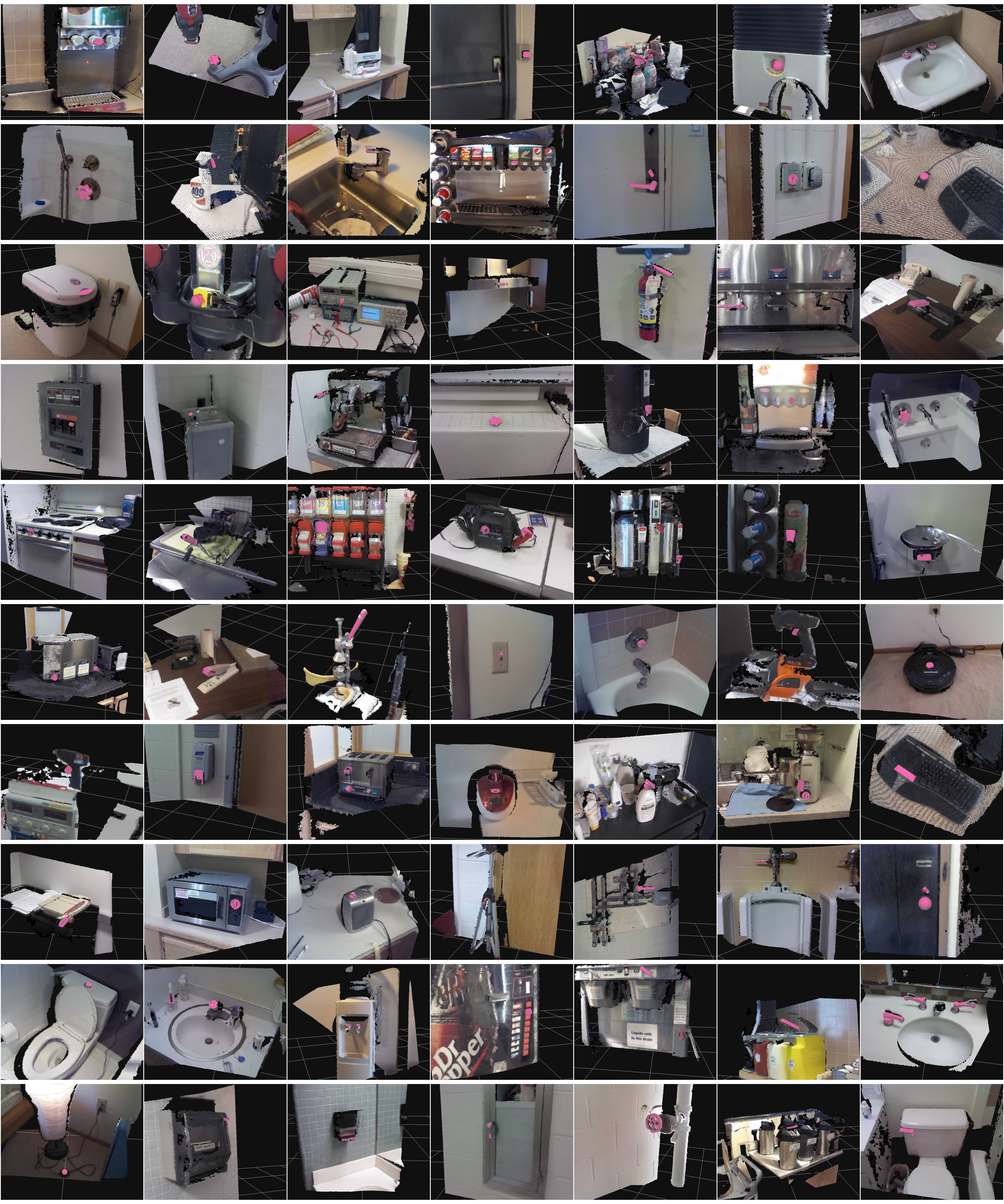}
\end{center}
\caption{
  \textbf{Examples of objects and object parts.} Each image shows the point cloud
  representation of an object. We overlaid some of its parts by CAD models for online Robobarista crowd-sourcing platform.
  Note that the actual underlying point-cloud of object parts contains much more noise and is not clearly segmented,
  and none of the models have access to overlaid model for inferring manipulation trajectory.
}
\label{fig:all_objects}
\end{figure*}

In order to test our model,
we have collected a dataset of 116 point-clouds of objects
with 249 object parts 
(examples shown in Figure~\ref{fig:data_ex}).
\jae{Objects range from kitchen appliances
such as stoves and rice cookers to bathroom hardware such as
sinks and toilets. Figure~\ref{fig:all_objects} shows a sample of 70 such objects.}
There are also a total of 250 natural language
instructions (in 155 manuals).\footnote{Although not necessary for training our model,
 we also collected trajectories from the expert for evaluation purposes.}
Using the crowd-sourcing platform Robobarista,
we collected 1225 trajectories for these objects
from 71 non-expert users on the Amazon Mechanical Turk.
After a user is shown a 20-second instructional video, the user first completes 
a 2-minute tutorial task.
At each session, the user was asked to complete 10 assignments 
where each consists of an object and a manual to be followed.

For each object, we took raw RGB-D images with the Microsoft Kinect sensor and
stitched them using Kinect Fusion \citep{izadi2011kinectfusion} to form a denser point-cloud
in order to incorporate different viewpoints of objects.
Objects range from kitchen appliances such as `stove', `toaster', and `rice cooker' to
`urinal', `soap dispenser', and `sink' in restrooms. 
The dataset is made available at {\small \url{http://robobarista.cs.cornell.edu}}


\begin{table*}[tb]
\begin{center}
\caption{
\textbf{Results on our dataset} with \emph{5-fold cross-validation}. 
Rows list models we tested including our model and baselines.
Columns show different metrics used to evaluate the models.
}
\begin{tabular}{r|c|c|c} 
\hline
  &  \textbf{per manual} & \multicolumn{2}{c}{\textbf{per instruction}} \\ \cline{2-4}
\textbf{Models}  &  \textbf{DTW-MT}  & \textbf{DTW-MT}  & \textbf{Accuracy ($\%$)} \\ \hline \hline
\emph{Chance}                  & $28.0 \;(\pm 0.8)$ & $27.8 \;(\pm 0.6)$ & $11.2 \;(\pm 1.0)$ \\ \hline
\emph{Object Part Classifier}  & - & $22.9 \;(\pm 2.2)$ & $23.3 \;(\pm 5.1)$ \\ 

\emph{Structured SVM}              & $21.0 \;(\pm 1.6)$ & $21.4 \;(\pm 1.6)$ & $26.9 \;(\pm 2.6)$ \\ 
\emph{Latent SSVM + Kinematic} \citep{sturm2011probabilistic}    & $17.4 \;(\pm 0.9)$ & $17.5 \;(\pm 1.6)$ & $40.8 \;(\pm 2.5)$\\ \hline

\emph{Task similarity + Random}                           & $14.4 \;(\pm 1.5)$ & $13.5 \;(\pm 1.4)$ & $49.4 \;(\pm 3.9)$   \\ 
\emph{Task Similarity + Weights} \citep{forbes2014robot}  & $13.3 \;(\pm 1.2)$ & $12.5 \;(\pm 1.2)$ & $53.7 \;(\pm 5.8)$     \\ \hline

\emph{Deep Network with Noise-handling without Embedding}     & $13.7 \;(\pm 1.6)$ &  $13.3 \;(\pm 1.6)$ &  $51.9 \;(\pm 7.9)$  \\ \hline

\emph{Deep Multimodal Network without Embedding}  & $14.0 \;(\pm 2.3)$ &  $13.7 \;(\pm 2.1)$ &  $49.7 \;(\pm 10.0)$    \\
\emph{Deep Multimodal Network with Noise-handling without Embedding} \cite{sung_robobarista_2015}      & $13.0 \;(\pm 1.3)$ &  $12.2 \;(\pm 1.1)$ &  $60.0 \;(\pm 5.1)$  \\ \hline

\emph{LMNN-like Cost Function} \citep{weinberger2005distance}      & $15.4 \;(\pm 1.8)$ &  $14.7 \;(\pm 1.6)$ &  $55.5 \;(\pm 5.3)$ \\ \hline    

\textit{Our Model without Any Pretraining}  & $13.2 \;(\pm 1.4)$ & $12.4 \;(\pm 1.0)$ & $54.2 \;(\pm 6.0)$  \\ \
\textit{Our Model with SDA}                 & $11.5 \;(\pm 0.6)$ &  $11.1 \;(\pm 0.6)$ &  $62.6 \;(\pm 5.8)$ \\ \hline

\jae{\textit{Our Model without Noise Handling}} & $12.6 \;(\pm 1.3)$ & $12.1 \;(\pm 1.1)$ & $53.8 \;(\pm 8.0)$ \\
\jae{\textit{Our Model with Experts}}           & $12.3 \;(\pm 0.5)$ & $11.8 \;(\pm 0.9)$ & $56.5 \;(\pm 4.5)$   \\ \hline

\textbf{\emph{Our Model - Deep Multimodal Embedding}}   & $\textbf{11.0}\;(\pm 0.8)$ & $\textbf{10.5}\;(\pm 0.7)$ & $\textbf{65.1}$\; $(\pm 4.9)$  \\ \hline  

\end{tabular}
\label{tab:results}
\end{center}
\end{table*}


\subsection{Baselines}
\label{sec:baselines}

We compared our model against \jae{many} baselines:

\noindent
1) \textit{Random Transfers (chance)}:
Trajectories are selected at random from the set of trajectories in the training set.

\noindent
2) \textit{Object Part Classifier}:
\jin{
To test our hypothesis that classifying object parts as an intermediate step does not
guarantee successful transfers, we trained an object part classifier using
multiclass SVM \citep{tsochantaridis2004support} on point-cloud features $\phi(p)$
including local shape features \citep{koppula2011semantic},
histogram of curvatures \citep{Rusu_ICRA2011_PCL}, and distribution of points.
Using this classifier, we first classify the target object part $p$ into an object part category (e.g. `handle', `knob'), then use the same feature space to 
find its nearest neighbor $p'$ of the same class from the training set. 
Then the trajectory $\tau'$ of $p'$ is transferred to $p$.
}




\noindent
3) \textit{Structured support vector machine (SSVM)}:
\jin{
We used SSVM to learn a discriminant scoring function 
$\mathcal{F} : \mathcal{P} \times \mathcal{L} \times \mathcal{T} \rightarrow \mathbb{R}$. 
At test time, for target point-cloud/language pair $(p, l)$,
we output the trajectory $\tau$ from the training set that
maximizes $\mathcal{F}$. To train SSVM, we use a joint feature mapping
$\phi(p, l, \tau) =  [\phi(\tau); \phi(p, \tau); \phi(l, \tau)]$. 
$\phi(\tau)$ applies Isomap \citep{tenenbaum2000global} to interpolated $\tau$
for non-linear dimensionality reduction. 
$\phi(p, \tau)$ captures the overall shape when trajectory $\tau$ is overlaid over 
point-cloud $p$ by jointly representing them in a voxel-based cube similar to Sec.~\ref{sec:preprocessing}, 
with each voxel holding count of occupancy by $p$ or $\tau$. Isomap is applied to this representation 
to get the final $\phi(p, \tau)$. 
Finally, $\phi(l, \tau)$ is the tensor product 
of the language features and trajectory features: $\phi(l, \tau) = \phi(l) \otimes \phi(\tau)$.
We used our loss function (Sec.~\ref{sec:metric}) to train SSVM and used 
the cutting plane method to solve the SSVM optimization problem \citep{joachims2009cutting}.
}
\noindent
4) \textit{Latent Structured SVM (LSSVM) + kinematic structure}:
\jin{
The way in which an object is manipulated largely depends on its internal structure --
whether it has a `revolute', `prismatic', or `fixed' joint. 
Instead of explicitly trying to learn this structure, we encoded this internal structure 
as latent variable $z \in \mathcal{Z}$, composed of joint type, center of joint, 
and axis of joint \citep{sturm2011probabilistic}. 
We used Latent SSVM \citep{yu2009learning} to train with $z$, learning the discriminant function 
$\mathcal{F} : \mathcal{P} \times \mathcal{L} \times \mathcal{T} \times \mathcal{Z} \rightarrow \mathbb{R}$.
The model was trained with feature mapping $\phi(p, l, \tau, z) = [\phi(\tau); \phi(p, \tau); \phi(l, \tau); \phi(l, z); \phi(p, \tau, z)]$, 
which includes additional features that involve $z$. 
$\phi(l, z)$ captures the relation between $l$, a bag-of-words representation of language, 
and bag-of-joint-types encoded by $z$ (vector of length 3 indicating existence of each joint type) 
by computing the tensor product $\phi(l) \otimes \phi(z)$, then reshaping the product into a vector.
$\phi(p, \tau, z)$ captures how well the portion of $\tau$ that actually interacts with $p$ abides 
by the internal structure $h$. $\phi(p, \tau, z)$ is a concatenation of three types of features, 
one for each joint type. For `revolute' type joints, it includes deviation of trajectory from plane of rotation defined by $z$, 
the maximum angular rotation while maintaining pre-defined proximity to the plane of rotation, 
and the average cosine similarity between rotation axis of $\tau$ and axis defined by $z$. 
For `prismatic' joints, it includes the average cosine similarity between the extension axis 
and the displacement vector between waypoints. Finally, for `fixed' joints, 
it includes whether the \textit{uninteracting} part of $\tau$ has collision with the background $p$ 
since it is important to approach the object from correct angle.
}



\noindent
5) \textit{Task-Similarity Transfers + random}:
\jin{
We compute the pairwise similarities between the test case $(p_{test}, l_{test})$ and each training example $(p_{train}, l_{train})$, then transfer a trajectory $\tau$ associated with the training example of highest similarity. Pairwise similarity is defined as a convex combination of the cosine similarity in bag-of-words representations of language and the average mutual point-wise distance of two point-clouds after a fixed number of iterations of the ICP \citep{besl1992method} algorithm. If there are multiple trajectories associated with $(p_{train}, l_{train})$ of highest similarity, the trajectory for transfer is selected randomly.
}

\noindent
6) \textit{Task-similarity Transfers + weighting}:
\jin{
The previous method is problematic when non-expert demonstrations for a single task $(p_{train}, l_{train})$ vary in quality. 
\citet{forbes2014robot} introduces a score function for weighting
demonstrations based on weighted distance to the ``seed'' (expert) demonstration.
Adapting to our scenario of not having any expert demonstrations,
we select $\tau$
that has the lowest average distance from all other demonstrations for the same task,
with each distance measured with our loss function (Sec.~\ref{sec:metric}.)
This is similar to our noise handling approach in Sec.~\ref{sec:noise}.
}
\noindent
7) \textit{Deep Network without Embedding}:
\ian{We train a deep neural network to learn a similar scoring function 
$\mathcal{F} : \mathcal{P} \times \mathcal{L} \times \mathcal{T} \rightarrow \mathbb{R}$ to that learned for SSVM above. 
This model discriminatively projects the combination of point-cloud, language, 
and trajectory features to a score which represents how well the trajectory matches that 
point-cloud/language combination. Note that this is much less efficient than our joint embedding approach,
as it must consider all combinations of a new point-cloud/language pair and every training trajectory
to perform inference, as opposed to our model which need only project this new pair to our joint
embedding space.}
This deep learning model concatenates all the input of three modalities
and learns three hidden layers before the final layer. 






\begin{figure}
  \begin{center}
    \includegraphics[width=.8\columnwidth,trim={5.5cm 0cm 5cm 8.5cm},clip]{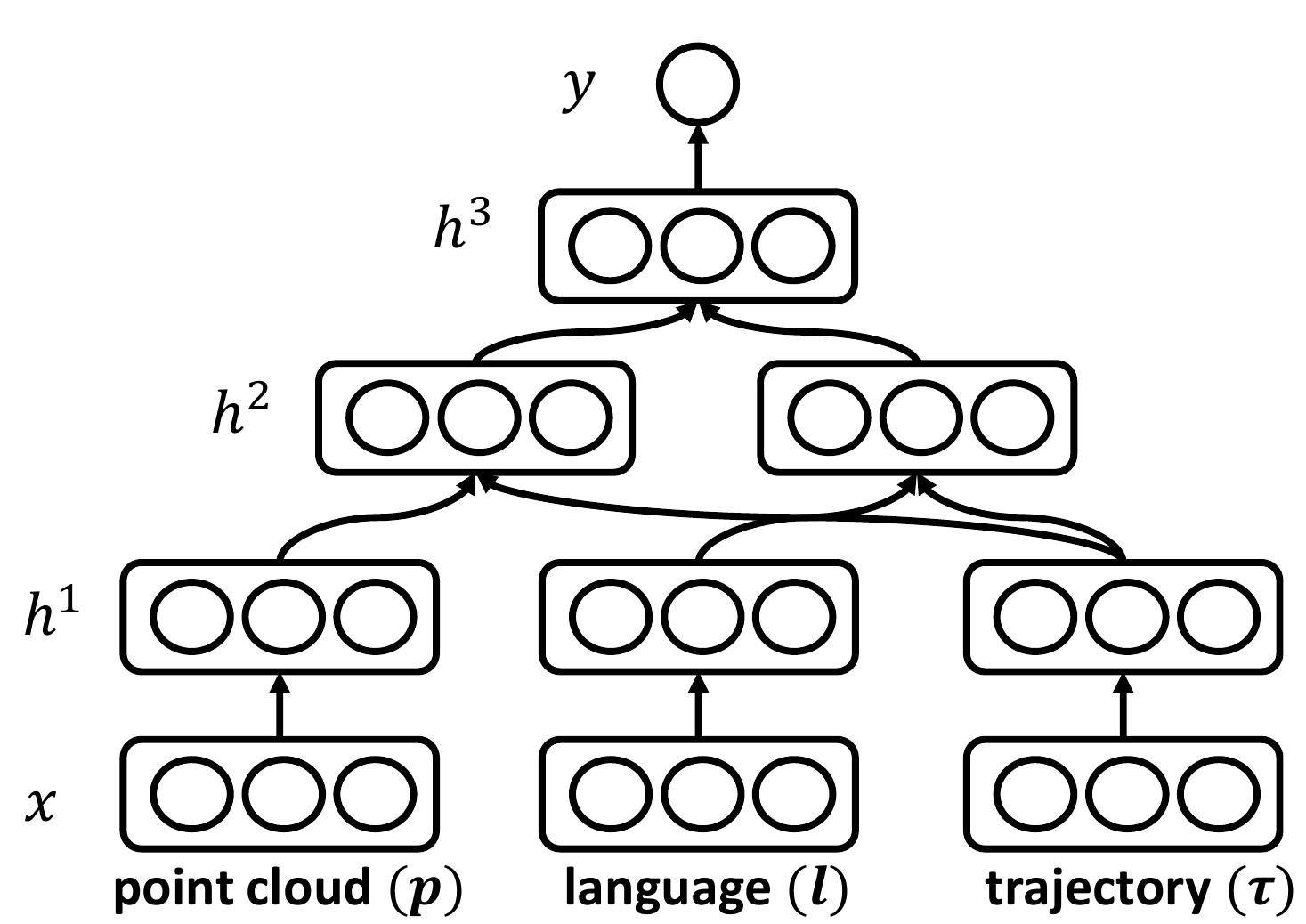}
  \end{center}
  \caption{
    \textbf{Deep Multimodal Network without Embedding} baseline model 
    takes the input $x$ of three different modalities (point-cloud, language, and trajectory)
    and outputs $y$, whether it is a good match or bad match. It first learns features separately ($h^1$) for each modality 
    and then learns the relation ($h^2$) between input and output of the original structured problem.
    Finally, last hidden layer $h^3$ learns relations of all these modalities.
    }
  \label{fig:isrr_model}
\end{figure}

\noindent
8) \textit{Deep Multimodal Network without Embedding}:
\jae{The same approach as `Deep Network without Embedding' with layers per each modality
before concatenating as shown in Figure~\ref{fig:isrr_model}.
More details about the model can be found in \cite{sung_robobarista_2015}.}


\begin{figure*}[tb]
\begin{center}
\includegraphics[width=\textwidth,trim={0cm 1cm 2cm 0cm}, clip]{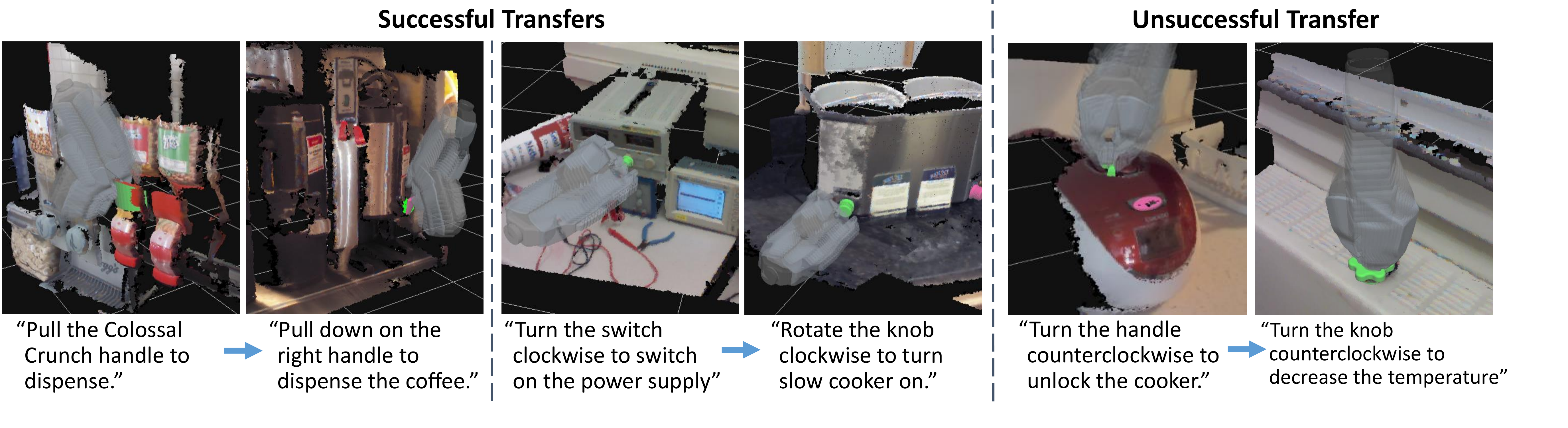}
\end{center}
\caption{
\textbf{Examples of successful and unsuccessful transfers} of manipulation trajectory from left to right
using our model.
In first two examples, though the robot has never seen  the `coffee dispenser' and `slow cooker' before, 
the robot has correctly identified that the trajectories of `cereal dispenser' and `DC power supply',
respectively, can be used to manipulate them. 
}
\label{fig:success_trans}
\end{figure*}

\noindent
9) \textit{LMNN \citep{weinberger2005distance}-like cost function:}
For all top layer fine-tuning and lower layer pre-training,
we define the cost function without loss augmentation.
Similar to LMNN \citep{weinberger2005distance}, 
we give a finite margin between similarities.
For example, as cost function for $h^3$:
\begin{align*}
L_{h^3}(p_i,l_i,\tau_i) = |1 + & sim(\Phi_{\mathcal{P},\mathcal{L}}(p_i,l_i), \Phi_{\mathcal{T}}(\tau'))  \\
- & sim(\Phi_{\mathcal{P},\mathcal{L}}(p_i,l_i), \Phi_{\mathcal{T}}(\tau_i))|_+
\end{align*}


\noindent
10) \textit{Our Model without Pretraining:}
Our full model finetuned without any pre-training of lower layers -- all parameters are
randomly initialized.

\noindent
11) \textit{Our Model with SDA:} 
	Our full model without pre-training $h^{2,pl}$ and $h^{2,\tau}$
	as defined in Section~\ref{sec:pre_task} and Section~\ref{sec:pre_traj}. 
Instead, we pre-train each layer as stacked de-noising autoencoders  \citep{vincent2008extracting,zeiler2013rectified}.

\noindent
12) \textit{Our Model without Noise Handling:}
Our model is trained without noise handling \ian{as presented in Section~\ref{sec:noise}}.
All of the trajectories collected from the crowd are trusted
as a ground-truth labels.

\noindent
13) \textit{Our Model with Experts:}
Our model is trained only using trajectory demonstrations from an expert which were collected
for evaluation purposes.

\noindent
14) \textit{Our Full Model - Deep Multimodal Embedding:}
\jae{Our full model as described in this paper with network size of
$h^{1,p}$, $h^{1,l}$, $h^{1,\tau}$, $h^{2,pl}$, $h^{2,\tau}$, and $h^3$
respectively having a layer with $150, 175, 100, 100, 75,$ and $50$ nodes.}



\subsection{Results and Discussions}
\label{sec:results}

We evaluated all models on our dataset using \emph{5-fold cross-validation}
and the results are in Table~\ref{tab:results}. 
All models which required hyper-parameter tuning used $10\%$ of the training data as the validation set.

Rows list the models we tested including our model and baselines. Each column
shows one of three evaluation metrics. The first two use dynamic time warping for
manipulation trajectory (DTW-MT) from Sec.~\ref{sec:metric}.
The first column shows averaged DTW-MT for each instruction manual 
consisting of one
or more language instructions. The second column shows averaged DTW-MT for every
test pair $(p,l)$.

\begin{figure}
  \begin{center}
    \includegraphics[width=0.95\columnwidth,trim={1.9cm 8cm 2cm 9.5cm},clip]{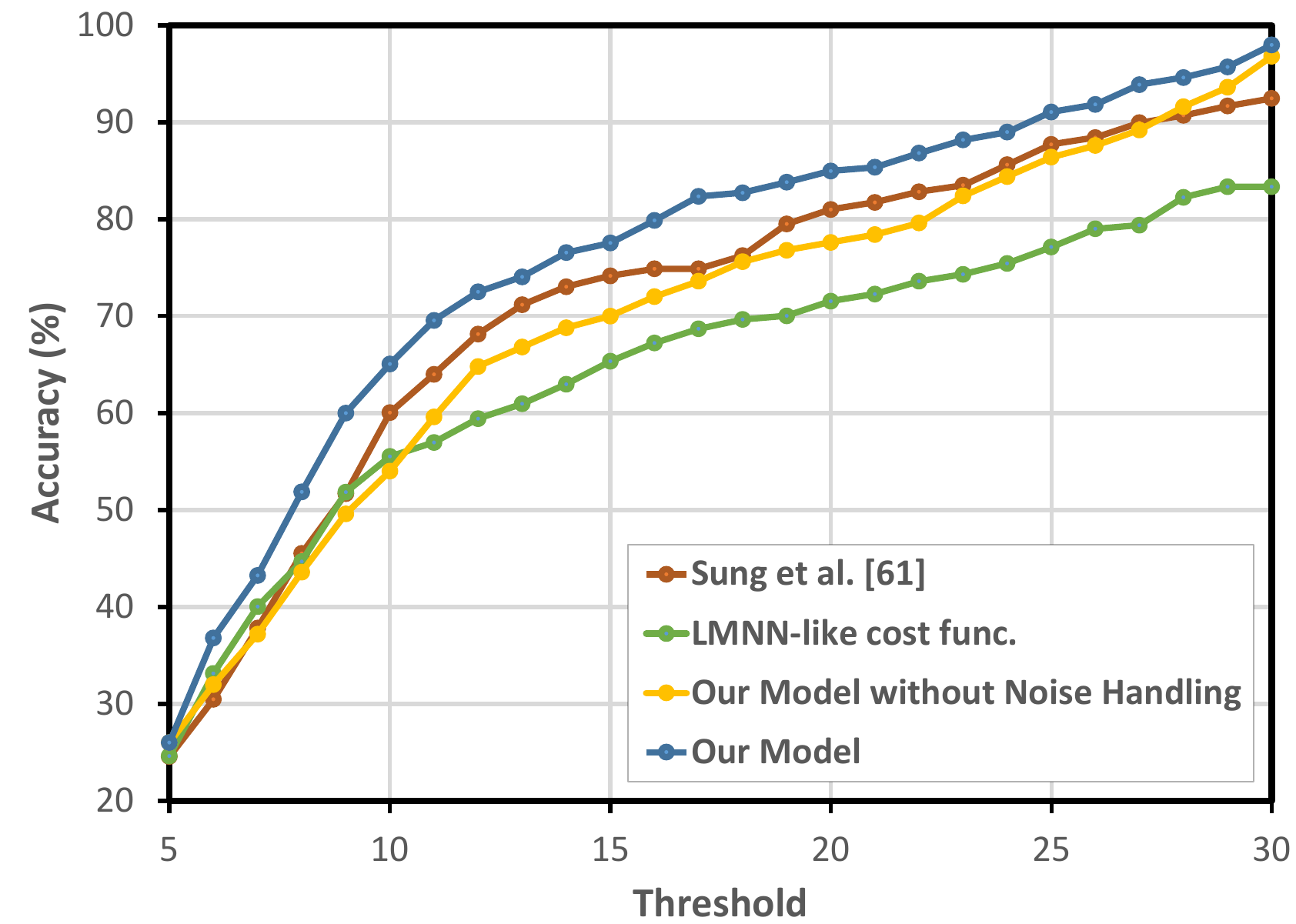}
  \end{center}
  \vskip -.15in
  \caption{
    \textbf{Thresholding Accuracy:} 
     Accuracy-threshold graph showing results of varying thresholds on DTW-MT scores. Our algorithm consistently
     outperforms the previous approach \citep{sung_robobarista_2015} and an
LMNN-like cost function \citep{weinberger2005distance}. 
     }
  \label{fig:accuracy_plot}
\end{figure}

As DTW-MT values are not intuitive, we also include a measure of ``accuracy,''
which shows the percentage of transferred trajectories with DTW-MT value less than $10$.
Through expert surveys, we found that when DTW-MT of manipulation trajectory
is less than $10$, 
the robot came up with a reasonable trajectory and will very likely be able to accomplish
the given task.
Additionally, Fig.~\ref{fig:accuracy_plot} shows accuracies obtained 
by varying the threshold on the DTW-MT measure.



\noindent
\textbf{Can manipulation trajectories be transferred from completely different objects?}
Our full model gave $65.1\%$ accuracy (Table~\ref{tab:results}),
outperforming every other baseline approach tested.

Fig.~\ref{fig:success_trans} shows two examples of successful transfers and one 
unsuccessful transfer by our model.
In the first example, the trajectory for pulling down on a cereal dispenser
is transferred to a coffee dispenser.
Because our approach to trajectory representation is based on the principal axis
 (Sec.~\ref{sec:transfer_adapt}),
even though the cereal and coffee dispenser handles are located and oriented differently, the transfer is a success.
The second example shows a successful transfer from a DC power supply to a slow
cooker, which have ``knobs'' of similar shape. The transfer was successful despite
the difference in instructions (``Turn the switch..'' and ``Rotate the knob..'')
and object type. This highlights the advantages of our end-to-end approach over relying
on semantic classes for parts and actions.

\jae{
The last example in Fig.~\ref{fig:success_trans} shows a potentially unsuccessful
transfer. Despite the similarity in two instructions and similarity in required counterclockwise motions, 
the transferred motion might not be successful.
While the knob on radiator must be grasped in the middle,
the rice cooker has a handle that extends sideways, requiring it to be grasped off-center.
For clarity of visualization in figures, 
we have overlaid CAD models over some noisy point-clouds.
Many of the object parts were too small and/or too glossy for the Kinect sensor.
We believe that a better 3-D sensor would allow for more accurate transfers. 
On the other hand, it is interesting to note that the transfer in opposite direction from the radiator knob to the rice cooker handle 
may have yielded a correct manipulation.
}

\begin{figure}[tb]
\begin{center}
\includegraphics[width=\columnwidth,trim={0cm 5cm 0cm 0cm}, clip]{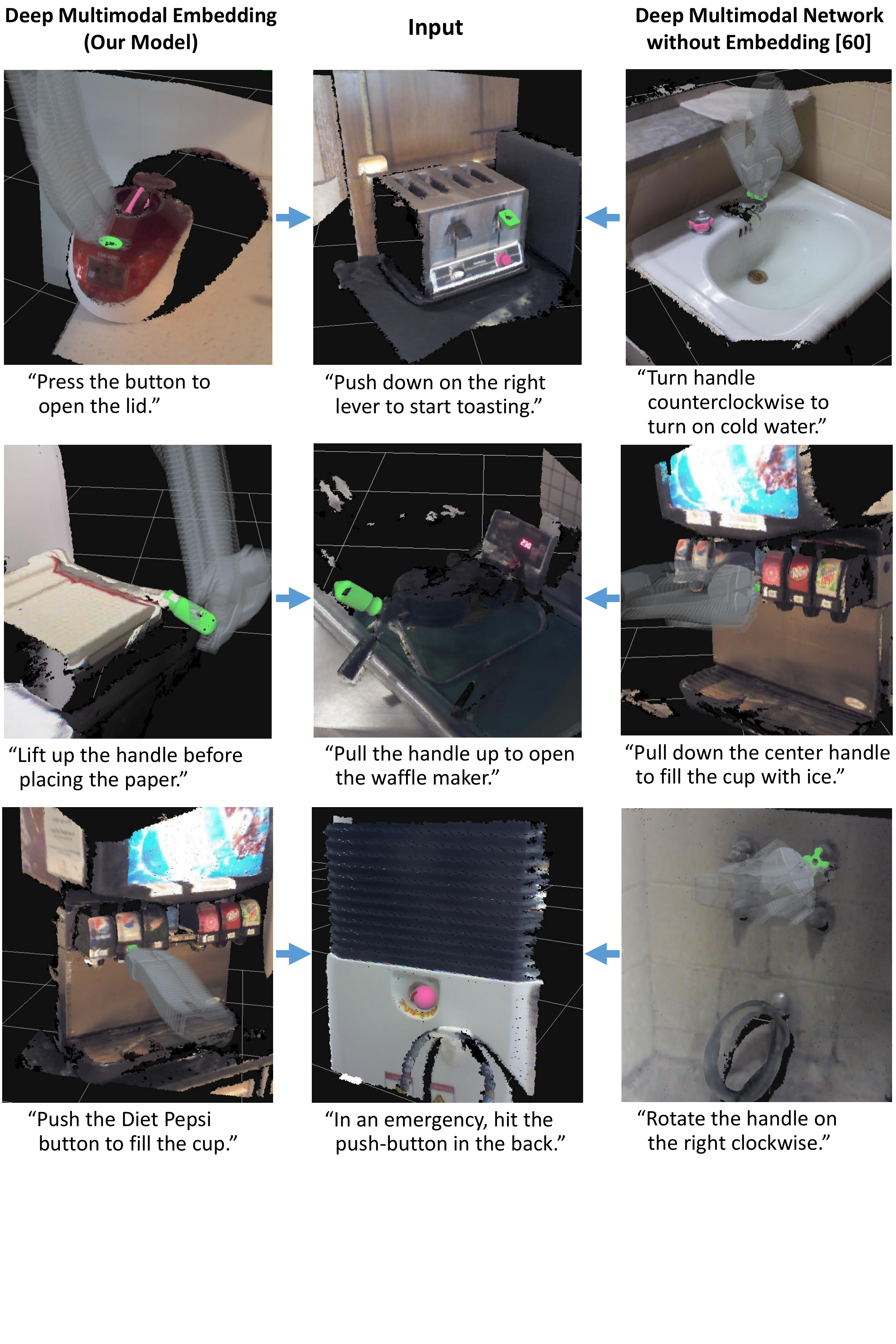}
\end{center}
\vskip -.1in
\caption{
\textbf{Comparisons of transfers} between our model and the baseline 
(deep multimodal network without embedding \cite{sung_robobarista_2015}).
In these three examples, our model successfully finds correct manipulation trajectory 
from these objects while the other one does not.
Given the lever of the toaster, our algorithm finds similarly slanted part from the rice cooker
while the other model finds completely irrelevant trajectory.
For the opening action of waffle maker, trajectory for paper cutter is correctly identified
while the other model transfers from a handle that has incompatible motion.
}
\label{fig:compare_transfer}
\end{figure}

\begin{figure*}[t]
\begin{center}
    \includegraphics[width=\textwidth,trim={0cm 0cm 0cm .5cm},clip]{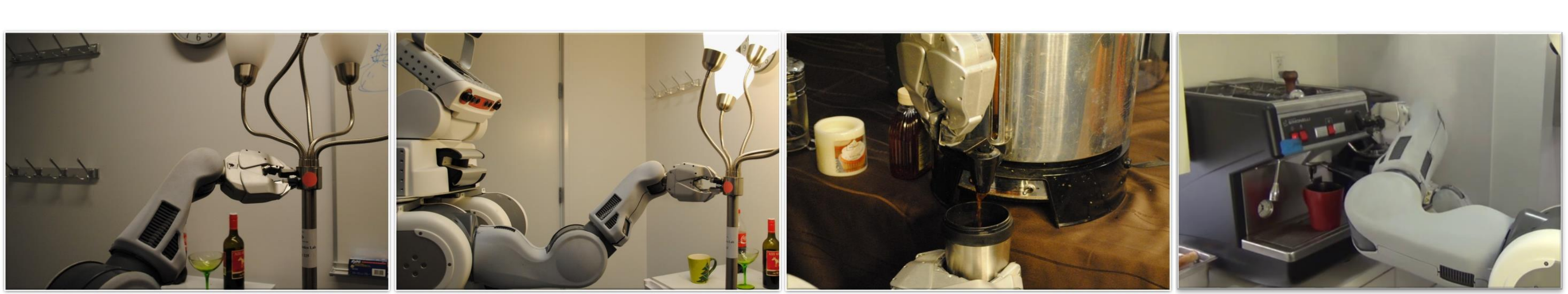}
\end{center}
\caption{
\textbf{Examples of transferred trajectories} being executed on PR2.
On the left, PR2 is able to rotate the `knob' to turn the lamp on.
In the third snapshot, using two transferred trajectories, PR2 is able to hold the cup below the `nozzle'
and press the `lever' of `coffee dispenser'.  
In the last example, PR2 is frothing milk by pulling down on the lever, and is able to prepare a cup of latte with many transferred trajectories.
}
\label{fig:robotic_exp}
\end{figure*}



\noindent
\textbf{Can we crowd-source the teaching of manipulation trajectories?}
When we trained our full model with expert demonstrations, which were collected for evaluation purposes, 
it performed at $56.5\%$ compared to $65.1\%$ by our model trained with crowd-sourced data.
Even though non-expert demonstrations can carry significant noise, as shown in last two examples of Fig.~\ref{fig:data_ex},
our noise-handling approach allowed our model to take advantage of the larger, less
accurate crowd-sourced dataset.
Note that all of our crowd users are true non-expert users from Amazon Mechanical Turk.

\noindent
\textbf{Is segmentation required for the system?}
Even with the state-of-the-art techniques \citep{felzenszwalb2010object,krizhevsky2012imagenet}, 
detection of `manipulable' object parts such as `handles' and 
`levers' in a point-cloud is by itself 
a challenging problem \citep{lai_icra14}.
Thus, we rely on human experts to pre-label parts of the object to be manipulated.
The point-cloud of the scene is over-segmented into thousands of supervoxels,
from which the expert chooses the part of the object to be manipulated.
Even with expert input, such segmented point-clouds are still extremely noisy
because of sensor failures, e.g. on glossy surfaces.

\noindent
\textbf{Is intermediate object part labeling necessary?}
A multiclass SVM trained on object part labels was able to obtain 
over $70\%$ recognition accuracy in 
classifying five major classes of object parts
(`button', `knob', `handle', `nozzle', `lever'.)
However, the \emph{Object Part Classifier} baseline, based on this classification,
performed at only $23.3\%$ accuracy for actual trajectory transfer, outperforming chance by merely 12.1\%, and
significantly underperforming our model's result of 65.1\%. This shows
that object part labels alone are not sufficient to enable manipulation motion
transfer, while our model, which makes use of richer information, does a much better
job.

\noindent
\textbf{Can features be hand-coded? What does learned deep embedding space represent?}
Even though we carefully designed state-of-the-art task-specific features for the
SSVM and LSSVM models, 
these models only gave at most 40.8\% accuracy.
The \textit{task similarity} method gave a better result of $53.7\%$, but 
it requires access to \emph{all} of the raw training data (point-clouds, language, and trajectories) at test time,
which leads to heavy computation at test time and requires a large amount of storage 
as the size of training data increases. Our approach, by contrast, requires only
the trajectory data, and a low-dimensional representation of the point-cloud
and language data, which is much less expensive to store than the raw data.



\begin{figure*}[tb]
  \begin{center}
    \includegraphics[width=\textwidth]{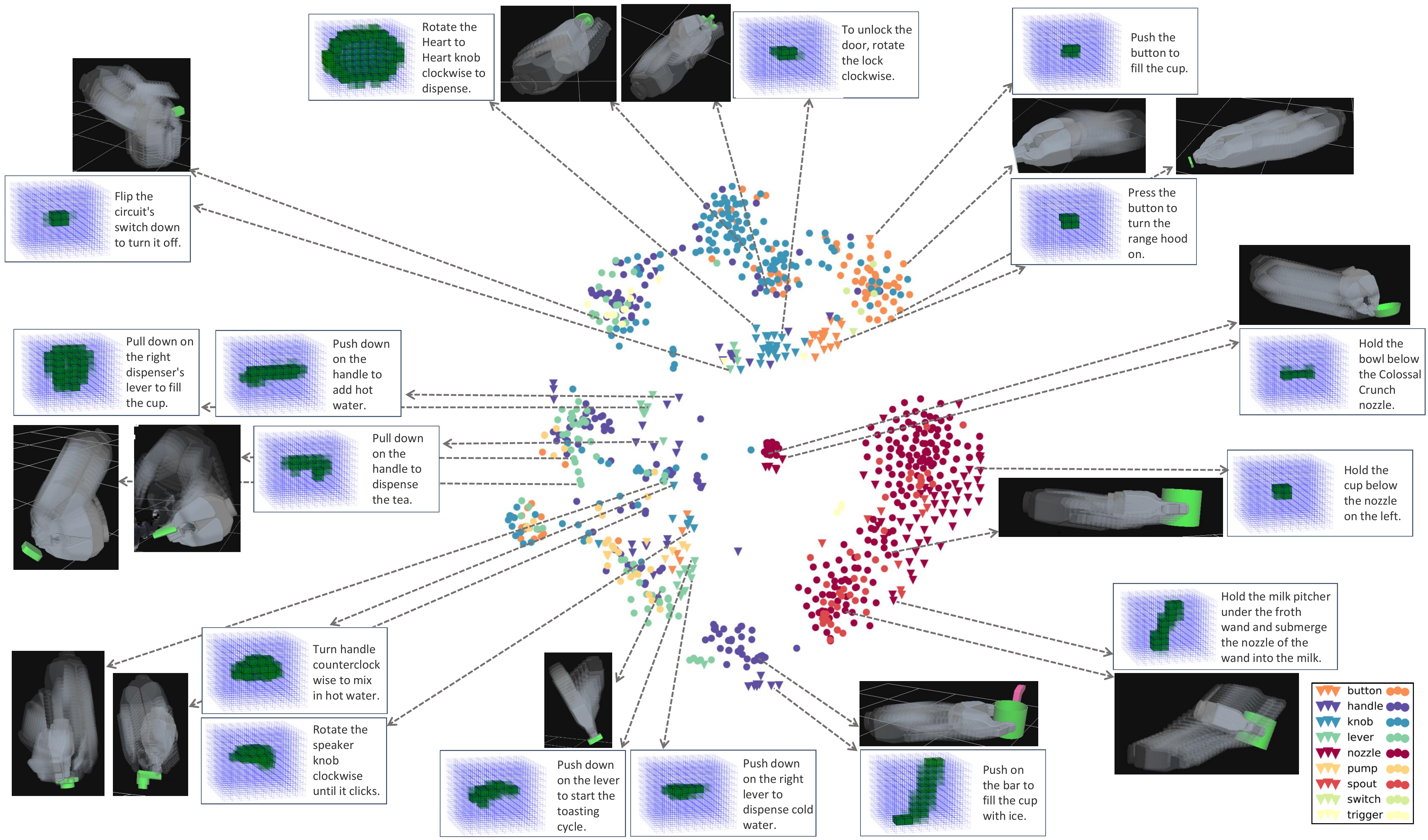}
  \end{center}
  \caption{
    \textbf{Learned Deep Point-cloud/Language/Trajectory Embedding Space: }
     Joint embedding space $h^3$ after the network is fully fine-tuned, 
     visualized in 2d using t-SNE \citep{van2008visualizing} .
     Inverted triangles represent projected point-cloud/language pairs, circles represent projected trajectories. 
     The occupancy grid representation of object part point-clouds is shown 
     in \textcolor{darkgreen}{green} in \textcolor{blue}{blue} grids.
     Among the two occupancy grids (Sec.~\ref{sec:preprocessing}), 
     we selected the one that is more visually parsable for each object.
     The legend at the bottom right shows classifications of object parts by an expert, 
     collected for the purpose of building a baseline.
     As shown by result of this baseline (object part classifier in Table~\ref{tab:results}), 
     these labels do not necessarily correlate well with the actual manipulation motion.
     Thus, full separation according to the labels defined in the legend is not optimal and
     will not occur in this figure or Fig.~\ref{fig:task_embed_vis}.
     These figures are best viewed in color. 
     }
  \label{fig:embed_vis}

\end{figure*}

\begin{figure*}[tb]
  \begin{center}
    \includegraphics[width=\textwidth]{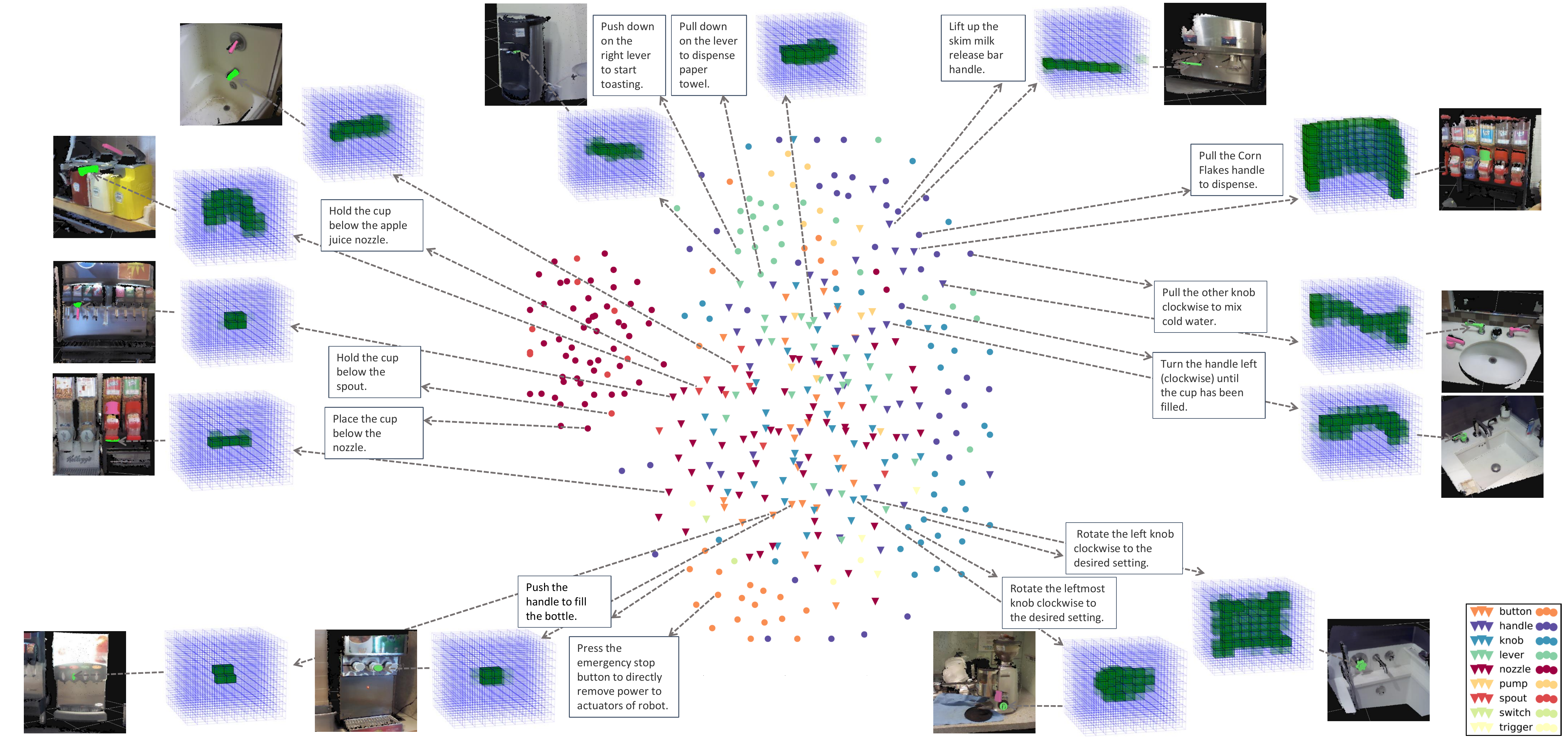}
  \end{center}
  \caption{ \textbf{Learned Point-cloud/Language Space: }
	  Visualization of the point-cloud/language layer $h^{2,lp}$ in 2d using t-SNE \citep{van2008visualizing} 
     after the network is fully fine-tuned.
     Inverted triangles represent projected point-clouds and circles represent projected instructions. 
     A subset of the embedded points are randomly selected for visualization.
     Since 3D point-clouds of object parts are hard to visualize, 
     we also include a snapshot of a point-cloud showing the whole object.
     Notice correlations in the motion required to manipulate the object or
     follow the instruction among nearby point-clouds and natural language.
    }
  \label{fig:task_embed_vis}
\end{figure*}

This shows that it is extremely difficult to find a good set of features which 
properly combines these three modalities.
Our multimodal embedding model does not require hand-designing such features,
instead learning a joint embedding space as shown by our visualization of the top layer $h^3$ in Figure~\ref{fig:embed_vis}.
This visualization is created by projecting all training data (point-cloud/language pairs and trajectories) 
of one of the cross-validation folds to $h^3$, 
then embedding them to 2-dimensional space 
using t-SNE \citep{van2008visualizing}.
Although previous work \citep{sung_robobarista_2015} was able to visualize several nodes in the top layer,
most were difficult to interpret.
With our model, we can embed all our data and 
visualize all the layers (see Figs.~\ref{fig:embed_vis} and \ref{fig:task_embed_vis}).

One interesting result is that our system was able to naturally learn that ``nozzle'' and ``spout''
are effectively synonyms for purposes of manipulation. It clustered these 
together in the lower-right of Fig.~\ref{fig:embed_vis} based solely on the
fact that both are associated with similar point-cloud shapes and manipulation
trajectories. 
At the same time, it also identified one exception, a small cluster of ``nozzles'' in the center
of Fig.~\ref{fig:embed_vis} which require
different manipulation motions.

In addition to the aforementioned cluster in the
bottom-right of Fig.~\ref{fig:embed_vis}, we see several
other logical clusters. 
Importantly, we can see that our embedding maps vertical
and horizontal rotation operations to very different 
regions of the space -- roughly 12 o'clock and 8 o'clock
in Fig.~\ref{fig:embed_vis}, respectively. Even though
these have nearly identical language instructions, our
algorithm learns to map them differently based on their
point-clouds, mapping nearby the appropriate manipulation
trajectories.

\noindent
\textbf{Should cost function be loss-augmented?}
When we changed the cost function for pre-training $h^2$ and fine-tuning $h^3$
to use a constant margin of $1$ between relevant $\mathcal{T}_{i,S}$ and 
irrelevant $\mathcal{T}_{i,D}$ demonstrations \citep{weinberger2005distance},
performance drops to $55.5\%$.
This loss-augmentation is also visible in our embedding space.
Notice the purple cluster around the 6 o'clock region of Fig.~\ref{fig:embed_vis},
and the lower part of the cluster in the 5 o'clock region.
The purple cluster represents tasks and demonstrations related to pushing a bar (often found on soda fountains),
and the lower part of the red cluster represents the task of holding a cup below the nozzle.
Although the motion required for one task would not be replaceable by the other,
the motions and shapes are very similar, especially compared to 
most other motions
e.g. turning a horizontal knob.

\noindent
\textbf{Is pre-embedding important?}
As seen in Table~\ref{tab:results}, without any pre-training
our model gives an accuracy of only $54.2\%$. Pre-training
the lower layers with the conventional stacked 
de-noising auto-encoder (SDA) algorithm \citep{vincent2008extracting,zeiler2013rectified}
increases performance to $62.6\%$, still significantly
underperforming our pre-training algorithm, 
which gives $65.1\%$. 
This shows that our metric embedding pre-training approach provides a better
initialization for an embedding space than SDA.

Fig.~\ref{fig:task_embed_vis} shows the 
joint point-cloud/language embedding $h^{2,pl}$
after the network is initialized using our pre-training
algorithm and then fine-tuned using our cost function
for $h^3$. 
While this space is not as
clearly clustered as $h^3$ shown in Fig.~\ref{fig:embed_vis}, 
we note that point-clouds tend to
appear in the more general center of the space, while natural language 
instructions appear around the more-specific edges. This makes sense because
one point-cloud might afford many possible actions, while language instructions
are much more specific.

\noindent
\textbf{Does embedding improve efficiency?}
The previous model \citep{sung_robobarista_2015} had $749,638$ parameters to be
learned, while our model has only 
$418,975$ (and still gives better performance.)

The previous model had to compute joint point-cloud/language/trajectory features 
for all combinations of the current point-cloud/language pair with \emph{each} candidate trajectory
(i.e. all trajectories in the training set) to infer an optimal trajectory. 
This is inefficient and does not scale well with the number of training
datapoints.
However, our model pre-computes the projection of all trajectories into $h^3$.
Inference in our model then requires only projecting
the new point-cloud/language combination to $h^3$ once and 
finding the trajectory with maximal similarity in this embedding.

In practice, this results in a significant improvement in efficiency, 
decreasing the average time to infer a trajectory
from $2.3206$ms to $0.0135$ms, a speed-up of about $171$x.
Time was measured on the same hardware, with a GPU (GeForce GTX Titan X), using the Theano library \citep{Bastien-Theano-2012}.
We measured inference times $10000$ times for first test fold, which has a pool of $962$ trajectories.
Time to preprocess the data and time to load into GPU memory was not included in this measurement.
We note that the only part of our algorithm's runtime
which scales up with the amount of training data is the nearest-neighbor
computation, for which there exist many efficient algorithms \citep{flann_pami_2014}.
Thus, our algorithm could be scaled to much larger datasets, allowing it to handle a wider
variety of tasks, environments, and objects.


\subsection{Robotic Validation} 
\label{sec:robotic_exp}




\ian{
To validate the concept of part-based manipulation trajectory transfer
in a real-world setting, we tested our algorithm on our PR2 robot. 
To ensure real transfers, we tested with four objects the algorithm had
never seen before -- a coffee dispenser, coffee grinder, lamp, and espresso 
machine.
}

\jae{
The PR2 robot has two 7DoF arms, an omni-directional base, and 
 many sensors including a Microsoft Kinect, stereo cameras, and a tilting laser scanner.
For these experiments, a point-cloud is acquired from the head mounted Kinect sensor and
each motion is executed on the specified arm
using a Cartesian end-effector stiffness controller \cite{bollini2011bakebot} in ROS \cite{quigley2009ros}.
}

\jae{
For each object, 
the robot is presented with a segmented point-cloud along with a natural language text manual, with each step in the manual
associated with a segmented part in the point-cloud.
Once our algorithm outputs a trajectory (transferred from a completely different
object), we find the manipulation frame for the part's point-cloud by using its
principal axis (Sec.~\ref{sec:preprocessing}).
Then, the transferred trajectory can be executed relative to the part using this
coordinate frame, without any modification to the trajectory.
}

\jae{
The chosen manipulation trajectory, defined as a set of waypoints, is
converted to a smooth and densely interpolated trajectory (Sec.~\ref{sec:prob_form}.)
The robot first computes and execute a collision-free motion 
to the starting point of the manipulation trajectory.
Then, starting from this first waypoint, the interpolated trajectory is executed. 
For these experiments, we placed the robot in reach of the object,
but one could also find a location using a motion planner that would make all 
waypoints of the manipulation trajectory reachable.
}

\jae{
Some of the examples of successful execution on a PR2 robot are shown in Fig.~\ref{fig:robotic_exp}
and in video at the project website: 
{\small \url{http://robobarista.cs.cornell.edu/}}.
For example, a manipulation trajectory from the task of ``turning on a light switch'' is transferred to 
the task of ``flipping on a switch to start extracting espresso'', 
and a trajectory for turning on DC power supply (by rotating clockwise) is transferred to turning on the floor lamp.
These demonstrations shows that part-based transfer of manipulation trajectories
is feasible without any modification to the source trajectories
by carefully choosing their representation and coordinate frames (Sec.~\ref{sec:transfer_adapt}).
}

%
%
%
%

\section{Conclusion and Future Work}

\jae{
In this work, we introduce a novel approach to predicting manipulation trajectories via part based transfer,
which allows robots to successfully manipulate even objects they have never seen before.
We formulate this as a structured-output problem and 
approach the problem of inferring manipulation trajectories
for novel objects by jointly embedding point-cloud, natural language, and trajectory data into a common space
using a deep neural network.
We introduce a method for learning a common representation of multimodal data
using a new loss-augmented cost function, which learns a semantically meaningful embedding from data.
We also introduce a method for pre-training the network's lower layers, learning embeddings for subsets of modalities, 
and show that it outperforms standard pre-training algorithms.
Learning such an embedding space allows efficient inference
by comparing the embedding of a new point-cloud/language pair 
against pre-embedded demonstrations.
We introduce our crowd-sourcing platform, Robobarista, 
which allows non-expert users to easily give manipulation demonstrations over the web.
This enables us to collect a large-scale dataset of 249 object parts with 1225 crowd-sourced demonstrations, on which our algorithm outperforms all other methods
tried.
We also verify on our robot that even manipulation trajectories transferred from completely different objects
can be used to successfully manipulate novel objects the robot has never seen before.
}

\ian{
While our model is able to give correct manipulation trajectories for most of the
objects we tested it on, outperforming all other approaches tried, 
open-loop execution of a pose trajectory may not be enough to correctly manipulate
some objects.
}
\jae{
For such objects, correctly executing a transferred manipulation trajectory may require incorporating visual
and/or force feedback \citep{wieland2009combining,vina2013predicting} 
in order for the execution to adapt exactly to the object.
For example, some dials or buttons have to be rotated or pushed until they click, 
and each might require a different amount of displacement to accomplish this task.}
\ian{For such tasks, the robot would have to use this feedback to adapt its 
trajectory in an online fashion.}

\ian{Our current model also only takes into account the object part and desired
action in question. For some objects, a correct trajectory according to these might
still collide with other parts of the environment. Once again, solving this problem
would require adapting the manipulation trajectory after it's selected, and is an
interesting direction for future work.}

\section*{ACKNOWLEDGMENT}
We thank Joshua Reichler for building the initial prototype of the crowd-sourcing
platform. 
We thank Ross Knepper and Emin G\"un Sirer for useful discussions. We thank
NVIDIA Corporation for the donation of the Tesla K40
GPU used for this research. This work was supported by
NRI award 1426452, ONR award N00014-14-1-0156, and
by Microsoft Faculty Fellowship and NSF Career Award to
one of us (Saxena).

{

\bibliography{writeup}

\begin{thebibliography}{78}
\providecommand{\natexlab}[1]{#1}
\providecommand{\url}[1]{\texttt{#1}}
\expandafter\ifx\csname urlstyle\endcsname\relax
  \providecommand{\doi}[1]{doi: #1}\else
  \providecommand{\doi}{doi: \begingroup \urlstyle{rm}\Url}\fi

\bibitem[Abbeel et~al.(2010)Abbeel, Coates, and Ng]{abbeel2010autonomous}
P.~Abbeel, A.~Coates, and A.~Ng.
\newblock Autonomous helicopter aerobatics through apprenticeship learning.
\newblock \emph{International Journal of Robotics Research}, 2010.

\bibitem[Aha et~al.(1991)Aha, Kibler, and Albert]{aha1991instance}
D.~W. Aha, D.~Kibler, and M.~K. Albert.
\newblock Instance-based learning algorithms.
\newblock \emph{Machine learning}, 1991.

\bibitem[Alexander et~al.(2012)Alexander, Hsiao, Jenkins, Suay, and
  Toris]{alexander2012robot}
B.~Alexander, K.~Hsiao, C.~Jenkins, B.~Suay, and R.~Toris.
\newblock Robot web tools [ros topics].
\newblock \emph{Robotics \& Automation Magazine, IEEE}, 19\penalty0
  (4):\penalty0 20--23, 2012.

\bibitem[Argall et~al.(2009)Argall, Chernova, Veloso, and
  Browning]{argall2009survey}
B.~Argall, S.~Chernova, M.~Veloso, and B.~Browning.
\newblock A survey of robot learning from demonstration.
\newblock \emph{RAS}, 2009.

\bibitem[Bastien et~al.(2012)Bastien, Lamblin, Pascanu,
  et~al.]{Bastien-Theano-2012}
F.~Bastien, P.~Lamblin, R.~Pascanu, et~al.
\newblock Theano: new features and speed improvements.
\newblock Deep Learning and Unsupervised Feature Learning NIPS 2012 Workshop,
  2012.

\bibitem[Bengio et~al.(2013)Bengio, Courville, and
  Vincent]{bengio2013representation}
Y.~Bengio, A.~Courville, and P.~Vincent.
\newblock Representation learning: A review and new perspectives.
\newblock \emph{Pattern Analysis and Machine Intelligence, IEEE Transactions
  on}, 35\penalty0 (8):\penalty0 1798--1828, 2013.

\bibitem[Besl and McKay(1992)]{besl1992method}
P.~J. Besl and N.~D. McKay.
\newblock Method for registration of 3-d shapes.
\newblock In \emph{Robotics-DL tentative}, pages 586--606. International
  Society for Optics and Photonics, 1992.

\bibitem[Blaschko and Lampert(2008)]{blaschko2008learning}
M.~Blaschko and C.~Lampert.
\newblock Learning to localize objects with structured output regression.
\newblock In \emph{ECCV}. 2008.

\bibitem[Bollini et~al.(2011)Bollini, Barry, and Rus]{bollini2011bakebot}
M.~Bollini, J.~Barry, and D.~Rus.
\newblock Bakebot: Baking cookies with the pr2.
\newblock In \emph{IEEE/RSJ International Conference on Intelligent Robots and
  Systems PR2 Workshop}, 2011.

\bibitem[Crick et~al.(2011)Crick, Osentoski, Jay, and Jenkins]{crick2011human}
C.~Crick, S.~Osentoski, G.~Jay, and O.~C. Jenkins.
\newblock Human and robot perception in large-scale learning from
  demonstration.
\newblock In \emph{HRI}. ACM, 2011.

\bibitem[Dang and Allen(2012)]{dang2012semantic}
H.~Dang and P.~K. Allen.
\newblock Semantic grasping: Planning robotic grasps functionally suitable for
  an object manipulation task.
\newblock In \emph{IEEE/RSJ International Conference on Intelligent Robots and
  Systems}, 2012.

\bibitem[Daniel et~al.(2012)Daniel, Neumann, and Peters]{daniel2012learning}
C.~Daniel, G.~Neumann, and J.~Peters.
\newblock Learning concurrent motor skills in versatile solution spaces.
\newblock In \emph{IEEE/RSJ International Conference on Intelligent Robots and
  Systems}. IEEE, 2012.

\bibitem[Detry et~al.(2013)Detry, Ek, Madry, and Kragic]{detry2013learning}
R.~Detry, C.~H. Ek, M.~Madry, and D.~Kragic.
\newblock Learning a dictionary of prototypical grasp-predicting parts from
  grasping experience.
\newblock In \emph{IEEE International Conference on Robotics and Automation},
  2013.

\bibitem[Endres et~al.(2013)Endres, Trinkle, and Burgard]{endres2013learning}
F.~Endres, J.~Trinkle, and W.~Burgard.
\newblock Learning the dynamics of doors for robotic manipulation.
\newblock In \emph{IEEE/RSJ International Conference on Intelligent Robots and
  Systems}, 2013.

\bibitem[Erdogan et~al.(2014)Erdogan, Yildirim, and
  Jacobs]{erdogan2014transfer}
G.~Erdogan, I.~Yildirim, and R.~A. Jacobs.
\newblock Transfer of object shape knowledge across visual and haptic
  modalities.
\newblock In \emph{Proceedings of the 36th Annual Conference of the Cognitive
  Science Society}, 2014.

\bibitem[Felzenszwalb et~al.(2010)Felzenszwalb, Girshick, McAllester, and
  Ramanan]{felzenszwalb2010object}
P.~F. Felzenszwalb, R.~B. Girshick, D.~McAllester, and D.~Ramanan.
\newblock Object detection with discriminatively trained part-based models.
\newblock \emph{PAMI}, 32\penalty0 (9):\penalty0 1627--1645, 2010.

\bibitem[Forbes et~al.(2014)Forbes, Chung, Cakmak, and Rao]{forbes2014robot}
M.~Forbes, M.~J.-Y. Chung, M.~Cakmak, and R.~P. Rao.
\newblock Robot programming by demonstration with crowdsourced action fixes.
\newblock In \emph{Second AAAI Conference on Human Computation and
  Crowdsourcing}, 2014.

\bibitem[Gibson(1986)]{gibson1986ecological}
J.~J. Gibson.
\newblock \emph{The ecological approach to visual perception}.
\newblock Psychology Press, 1986.

\bibitem[Girshick et~al.(2011)Girshick, Felzenszwalb, and
  McAllester]{girshick2011object}
R.~Girshick, P.~Felzenszwalb, and D.~McAllester.
\newblock Object detection with grammar models.
\newblock In \emph{NIPS}, 2011.

\bibitem[Goodfellow et~al.(2009)Goodfellow, Le, Saxe, Lee, and Ng]{SAE}
I.~Goodfellow, Q.~Le, A.~Saxe, H.~Lee, and A.~Y. Ng.
\newblock Measuring invariances in deep networks.
\newblock In \emph{NIPS}, 2009.

\bibitem[Hadsell et~al.(2008)Hadsell, Erkan, Sermanet, Scoffier, Muller, and
  LeCun]{hadsell2008deep}
R.~Hadsell, A.~Erkan, P.~Sermanet, M.~Scoffier, U.~Muller, and Y.~LeCun.
\newblock Deep belief net learning in a long-range vision system for autonomous
  off-road driving.
\newblock In \emph{IEEE/RSJ International Conference on Intelligent Robots and
  Systems}, pages 628--633. IEEE, 2008.

\bibitem[Hinton and Salakhutdinov(2006)]{hinton2006reducing}
G.~Hinton and R.~Salakhutdinov.
\newblock Reducing the dimensionality of data with neural networks.
\newblock \emph{Science}, 313\penalty0 (5786):\penalty0 504--507, 2006.

\bibitem[Hsiao et~al.(2010)Hsiao, Chitta, Ciocarlie, and
  Jones]{hsiao2010contact}
K.~Hsiao, S.~Chitta, M.~Ciocarlie, and E.~Jones.
\newblock Contact-reactive grasping of objects with partial shape information.
\newblock In \emph{IEEE/RSJ International Conference on Intelligent Robots and
  Systems}, 2010.

\bibitem[Hu et~al.(2014{\natexlab{a}})Hu, Lu, and Tan]{hu2014discriminative}
J.~Hu, J.~Lu, and Y.-P. Tan.
\newblock Discriminative deep metric learning for face verification in the
  wild.
\newblock In \emph{The IEEE Conference on Computer Vision and Pattern
  Recognition}, 2014{\natexlab{a}}.

\bibitem[Hu et~al.(2014{\natexlab{b}})Hu, Lou, Englebienne, and
  Kröse]{Hu2014humanact}
N.~Hu, Z.~Lou, G.~Englebienne, and B.~Kröse.
\newblock Learning to recognize human activities from soft labeled data.
\newblock In \emph{Proceedings of Robotics: Science and Systems}, Berkeley,
  USA, July 2014{\natexlab{b}}.

\bibitem[Izadi et~al.(2011)Izadi, Kim, Hilliges, Molyneaux, Newcombe, Kohli,
  et~al.]{izadi2011kinectfusion}
S.~Izadi, D.~Kim, O.~Hilliges, D.~Molyneaux, R.~Newcombe, P.~Kohli, et~al.
\newblock Kinectfusion: real-time 3d reconstruction and interaction using a
  moving depth camera.
\newblock In \emph{ACM Symposium on UIST}, 2011.

\bibitem[Jain et~al.(2015)Jain, Wojcik, Joachims, and
  Saxena]{jainsaxena2015_learningpreferencesmanipulation}
A.~Jain, B.~Wojcik, T.~Joachims, and A.~Saxena.
\newblock Learning preferences for manipulation tasks from online coactive
  feedback.
\newblock In \emph{International Journal of Robotics Research}, 2015.

\bibitem[Joachims et~al.(2009)Joachims, Finley, and Yu]{joachims2009cutting}
T.~Joachims, T.~Finley, and C.-N.~J. Yu.
\newblock Cutting-plane training of structural svms.
\newblock \emph{Machine Learning}, 2009.

\bibitem[Katz et~al.(2013)Katz, Kazemi, Andrew~Bagnell, and
  Stentz]{katz2013interactive}
D.~Katz, M.~Kazemi, J.~Andrew~Bagnell, and A.~Stentz.
\newblock Interactive segmentation, tracking, and kinematic modeling of unknown
  3d articulated objects.
\newblock In \emph{IEEE International Conference on Robotics and Automation},
  pages 5003--5010. IEEE, 2013.

\bibitem[Kehoe et~al.(2013)Kehoe, Matsukawa, Candido, Kuffner, and
  Goldberg]{kehoe2013cloud}
B.~Kehoe, A.~Matsukawa, S.~Candido, J.~Kuffner, and K.~Goldberg.
\newblock Cloud-based robot grasping with the google object recognition engine.
\newblock In \emph{IEEE International Conference on Robotics and Automation},
  2013.

\bibitem[Koppula and Saxena(2013)]{koppula2013_anticipatingactivities}
H.~Koppula and A.~Saxena.
\newblock Anticipating human activities using object affordances for reactive
  robotic response.
\newblock In \emph{Robotics: Science and Systems}, 2013.

\bibitem[Koppula et~al.(2011)Koppula, Anand, Joachims, and
  Saxena]{koppula2011semantic}
H.~Koppula, A.~Anand, T.~Joachims, and A.~Saxena.
\newblock Semantic labeling of 3d point clouds for indoor scenes.
\newblock \emph{NIPS}, 2011.

\bibitem[Krizhevsky et~al.(2012)Krizhevsky, Sutskever, and
  Hinton]{krizhevsky2012imagenet}
A.~Krizhevsky, I.~Sutskever, and G.~E. Hinton.
\newblock Imagenet classification with deep convolutional neural networks.
\newblock In \emph{NIPS}, 2012.

\bibitem[Kroemer et~al.(2012)Kroemer, Ugur, Oztop, and
  Peters]{kroemer2012kernel}
O.~Kroemer, E.~Ugur, E.~Oztop, and J.~Peters.
\newblock A kernel-based approach to direct action perception.
\newblock In \emph{IEEE International Conference on Robotics and Automation},
  2012.

\bibitem[Lai et~al.(2014)Lai, Bo, and Fox]{lai_icra14}
K.~Lai, L.~Bo, and D.~Fox.
\newblock Unsupervised feature learning for 3d scene labeling.
\newblock In \emph{IEEE International Conference on Robotics and Automation},
  2014.

\bibitem[Lenz et~al.(2013)Lenz, Lee, and Saxena]{lenz2013deep}
I.~Lenz, H.~Lee, and A.~Saxena.
\newblock Deep learning for detecting robotic grasps.
\newblock \emph{Robotics: Science and Systems}, 2013.

\bibitem[Lenz et~al.(2015)Lenz, Knepper, and Saxena]{lenz2015deepmpc}
I.~Lenz, R.~Knepper, and A.~Saxena.
\newblock Deepmpc: Learning deep latent features for model predictive control.
\newblock In \emph{Robotics: Science and Systems}, 2015.

\bibitem[Levine et~al.(2015)Levine, Wagener, and
  Abbeel]{levine2015manipulation}
S.~Levine, N.~Wagener, and P.~Abbeel.
\newblock Learning contact-rich manipulation skills with guided policy search.
\newblock \emph{IEEE International Conference on Robotics and Automation},
  2015.

\bibitem[Li et~al.(2009)Li, Socher, and Fei-Fei]{li2009towards}
L.-J. Li, R.~Socher, and L.~Fei-Fei.
\newblock Towards total scene understanding: Classification, annotation and
  segmentation in an automatic framework.
\newblock In \emph{The IEEE Conference on Computer Vision and Pattern
  Recognition}, 2009.

\bibitem[Mangin et~al.(2011)Mangin, Oudeyer, et~al.]{mangin2011unsupervised}
O.~Mangin, P.-Y. Oudeyer, et~al.
\newblock Unsupervised learning of simultaneous motor primitives through
  imitation.
\newblock In \emph{IEEE ICDL-EPIROB}, 2011.

\bibitem[Mikolov et~al.(2013)Mikolov, Le, and
  Sutskever]{mikolov2013translation}
T.~Mikolov, Q.~V. Le, and I.~Sutskever.
\newblock Exploiting similarities among languages for machine translation.
\newblock \emph{CoRR}, 2013.

\bibitem[Miller et~al.(2012)Miller, Van Den~Berg, Fritz, Darrell, Goldberg, and
  Abbeel]{miller2012geometric}
S.~Miller, J.~Van Den~Berg, M.~Fritz, T.~Darrell, K.~Goldberg, and P.~Abbeel.
\newblock A geometric approach to robotic laundry folding.
\newblock \emph{International Journal of Robotics Research}, 2012.

\bibitem[Misra et~al.(2014)Misra, Sung, Lee, and Saxena]{misra2014tellme}
D.~Misra, J.~Sung, K.~Lee, and A.~Saxena.
\newblock Tell me dave: Context-sensitive grounding of natural language to
  mobile manipulation instructions.
\newblock In \emph{Robotics: Science and Systems}, 2014.

\bibitem[Moore et~al.(2012)Moore, Chen, Joachims, and
  Turnbull]{moore2012playlist}
J.~Moore, S.~Chen, T.~Joachims, and D.~Turnbull.
\newblock Learning to embed songs and tags for playlist prediction.
\newblock In \emph{Conference of the International Society for Music
  Information Retrieval (ISMIR)}, pages 349--354, 2012.

\bibitem[Muja and Lowe(2014)]{flann_pami_2014}
M.~Muja and D.~G. Lowe.
\newblock Scalable nearest neighbor algorithms for high dimensional data.
\newblock \emph{PAMI}, 2014.

\bibitem[M{\"u}lling et~al.(2013)M{\"u}lling, Kober, Kroemer, and
  Peters]{mulling2013learning}
K.~M{\"u}lling, J.~Kober, O.~Kroemer, and J.~Peters.
\newblock Learning to select and generalize striking movements in robot table
  tennis.
\newblock \emph{International Journal of Robotics Research}, 32\penalty0
  (3):\penalty0 263--279, 2013.

\bibitem[Ngiam et~al.(2011)Ngiam, Khosla, Kim, Nam, Lee, and
  Ng]{ngiam2011multimodal}
J.~Ngiam, A.~Khosla, M.~Kim, J.~Nam, H.~Lee, and A.~Y. Ng.
\newblock Multimodal deep learning.
\newblock In \emph{International Conference on Machine Learning}, 2011.

\bibitem[Pastor et~al.(2009)Pastor, Hoffmann, Asfour, and
  Schaal]{pastor2009learning}
P.~Pastor, H.~Hoffmann, T.~Asfour, and S.~Schaal.
\newblock Learning and generalization of motor skills by learning from
  demonstration.
\newblock In \emph{IEEE International Conference on Robotics and Automation},
  2009.

\bibitem[Phillips et~al.(2013)Phillips, Hwang, Chitta, and
  Likhachev]{phillipslearning}
M.~Phillips, V.~Hwang, S.~Chitta, and M.~Likhachev.
\newblock Learning to plan for constrained manipulation from demonstrations.
\newblock In \emph{Robotics: Science and Systems}, 2013.

\bibitem[Pillai et~al.(2014)Pillai, Walter, and Teller]{pillai2014articulated}
S.~Pillai, M.~Walter, and S.~Teller.
\newblock Learning articulated motions from visual demonstration.
\newblock In \emph{Robotics: Science and Systems}, 2014.

\bibitem[Quigley et~al.(2009)Quigley, Conley, Gerkey, Faust, Foote, Leibs,
  Wheeler, and Ng]{quigley2009ros}
M.~Quigley, K.~Conley, B.~Gerkey, J.~Faust, T.~Foote, J.~Leibs, R.~Wheeler, and
  A.~Y. Ng.
\newblock Ros: an open-source robot operating system.
\newblock In \emph{ICRA workshop on open source software}, volume~3, page~5,
  2009.

\bibitem[Rusu and Cousins(2011)]{Rusu_ICRA2011_PCL}
R.~Rusu and S.~Cousins.
\newblock {3D is here: Point Cloud Library (PCL)}.
\newblock In \emph{IEEE International Conference on Robotics and Automation},
  2011.

\bibitem[Shoemake(1985)]{shoemake1985animating}
K.~Shoemake.
\newblock Animating rotation with quaternion curves.
\newblock \emph{SIGGRAPH}, 19\penalty0 (3):\penalty0 245--254, 1985.

\bibitem[Socher et~al.(2011)Socher, Pennington, Huang, Ng, and
  Manning]{socher2011semi}
R.~Socher, J.~Pennington, E.~Huang, A.~Ng, and C.~Manning.
\newblock Semi-supervised recursive autoencoders for predicting sentiment
  distributions.
\newblock In \emph{EMNLP}, 2011.

\bibitem[Socher et~al.(2012)Socher, Huval, Bhat, Manning, and
  Ng]{socher2012convolutional}
R.~Socher, B.~Huval, B.~Bhat, C.~Manning, and A.~Ng.
\newblock Convolutional-recursive deep learning for 3d object classification.
\newblock In \emph{NIPS}, 2012.

\bibitem[Sohn et~al.(2014)Sohn, Shang, and Lee]{sohn2014improved}
K.~Sohn, W.~Shang, and H.~Lee.
\newblock Improved multimodal deep learning with variation of information.
\newblock In \emph{NIPS}, 2014.

\bibitem[Srivastava and Salakhutdinov(2012)]{srivastava2012multimodal}
N.~Srivastava and R.~R. Salakhutdinov.
\newblock Multimodal learning with deep boltzmann machines.
\newblock In \emph{NIPS}, 2012.

\bibitem[Sturm et~al.(2011)Sturm, Stachniss, and
  Burgard]{sturm2011probabilistic}
J.~Sturm, C.~Stachniss, and W.~Burgard.
\newblock A probabilistic framework for learning kinematic models of
  articulated objects.
\newblock \emph{Journal of Artificial Intelligence Research}, 41\penalty0
  (2):\penalty0 477--526, 2011.

\bibitem[Sung et~al.(2012)Sung, Ponce, Selman, and
  Saxena]{sung_rgbdactivity_2012}
J.~Sung, C.~Ponce, B.~Selman, and A.~Saxena.
\newblock Unstructured human activity detection from rgbd images.
\newblock In \emph{IEEE International Conference on Robotics and Automation},
  2012.

\bibitem[Sung et~al.(2014)Sung, Selman, and
  Saxena]{sung_synthesizingsequences_2014}
J.~Sung, B.~Selman, and A.~Saxena.
\newblock Synthesizing manipulation sequences for under-specified tasks using
  unrolled markov random fields.
\newblock In \emph{IEEE/RSJ International Conference on Intelligent Robots and
  Systems}, 2014.

\bibitem[Sung et~al.(2015)Sung, Jin, and Saxena]{sung_robobarista_2015}
J.~Sung, S.~H. Jin, and A.~Saxena.
\newblock Robobarista: Object part-based transfer of manipulation trajectories
  from crowd-sourcing in 3d pointclouds.
\newblock In \emph{International Symposium on Robotics Research (ISRR)}, 2015.

\bibitem[Tellex et~al.(2014)Tellex, Knepper, Li, Howard, Rus, and
  Roy]{tellex2014asking}
S.~Tellex, R.~Knepper, A.~Li, T.~Howard, D.~Rus, and N.~Roy.
\newblock Asking for help using inverse semantics.
\newblock \emph{Robotics: Science and Systems}, 2014.

\bibitem[Tenenbaum et~al.(2000)Tenenbaum, De~Silva, and
  Langford]{tenenbaum2000global}
J.~Tenenbaum, V.~De~Silva, and J.~Langford.
\newblock A global geometric framework for nonlinear dimensionality reduction.
\newblock \emph{Science}, 290\penalty0 (5500):\penalty0 2319--2323, 2000.

\bibitem[Thrun et~al.(2005)Thrun, Burgard, Fox, et~al.]{thrun2005probabilistic}
S.~Thrun, W.~Burgard, D.~Fox, et~al.
\newblock \emph{Probabilistic robotics}.
\newblock MIT press Cambridge, 2005.

\bibitem[Toris and Chernova(2013)]{toris2013robotsfor}
R.~Toris and S.~Chernova.
\newblock Robotsfor. me and robots for you.
\newblock In \emph{Proceedings of the Interactive Machine Learning Workshop,
  Intelligent User Interfaces Conference}, pages 10--12, 2013.

\bibitem[Toris et~al.(2014)Toris, Kent, and Chernova]{toris2014robot}
R.~Toris, D.~Kent, and S.~Chernova.
\newblock The robot management system: A framework for conducting human-robot
  interaction studies through crowdsourcing.
\newblock \emph{Journal of Human-Robot Interaction}, 3\penalty0 (2):\penalty0
  25--49, 2014.

\bibitem[Tsochantaridis et~al.(2004)Tsochantaridis, Hofmann, Joachims, and
  Altun]{tsochantaridis2004support}
I.~Tsochantaridis, T.~Hofmann, T.~Joachims, and Y.~Altun.
\newblock Support vector machine learning for interdependent and structured
  output spaces.
\newblock In \emph{ICML}. ACM, 2004.

\bibitem[Tsochantaridis et~al.(2005)Tsochantaridis, Joachims, Hofmann, Altun,
  and Singer]{tsochantaridis2005large}
I.~Tsochantaridis, T.~Joachims, T.~Hofmann, Y.~Altun, and Y.~Singer.
\newblock Large margin methods for structured and interdependent output
  variables.
\newblock \emph{JMLR}, 6\penalty0 (9), 2005.

\bibitem[Van~der Maaten and Hinton(2008)]{van2008visualizing}
L.~Van~der Maaten and G.~Hinton.
\newblock Visualizing data using t-sne.
\newblock \emph{Journal of Machine Learning Research}, 9\penalty0
  (2579-2605):\penalty0 85, 2008.

\bibitem[Vina et~al.(2013)Vina, Bekiroglu, Smith, Karayiannidis, and
  Kragic]{vina2013predicting}
F.~Vina, Y.~Bekiroglu, C.~Smith, Y.~Karayiannidis, and D.~Kragic.
\newblock Predicting slippage and learning manipulation affordances through
  gaussian process regression.
\newblock In \emph{Humanoids}, 2013.

\bibitem[Vincent et~al.(2008)Vincent, Larochelle, Bengio, and
  Manzagol]{vincent2008extracting}
P.~Vincent, H.~Larochelle, Y.~Bengio, and P.-A. Manzagol.
\newblock Extracting and composing robust features with denoising autoencoders.
\newblock In \emph{International Conference on Machine Learning}, 2008.

\bibitem[Weinberger et~al.(2005)Weinberger, Blitzer, and
  Saul]{weinberger2005distance}
K.~Q. Weinberger, J.~Blitzer, and L.~K. Saul.
\newblock Distance metric learning for large margin nearest neighbor
  classification.
\newblock In \emph{NIPS}, 2005.

\bibitem[Weston et~al.(2011)Weston, Bengio, and Usunier]{weston2011wsabie}
J.~Weston, S.~Bengio, and N.~Usunier.
\newblock Wsabie: Scaling up to large vocabulary image annotation.
\newblock In \emph{IJCAI}, 2011.

\bibitem[Wieland et~al.(2009)Wieland, Gonzalez-Aguirre, Vahrenkamp, Asfour, and
  Dillmann]{wieland2009combining}
S.~Wieland, D.~Gonzalez-Aguirre, N.~Vahrenkamp, T.~Asfour, and R.~Dillmann.
\newblock Combining force and visual feedback for physical interaction tasks in
  humanoid robots.
\newblock In \emph{Humanoid Robots}, 2009.

\bibitem[Wu et~al.(2014)Wu, Lenz, and Saxena]{wu2014_hierarchicalrgbd}
C.~Wu, I.~Lenz, and A.~Saxena.
\newblock Hierarchical semantic labeling for task-relevant rgb-d perception.
\newblock In \emph{Robotics: Science and Systems}, 2014.

\bibitem[Yu and Joachims(2009)]{yu2009learning}
C.-N. Yu and T.~Joachims.
\newblock Learning structural svms with latent variables.
\newblock In \emph{International Conference on Machine Learning}, 2009.

\bibitem[Zeiler(2012)]{zeiler2012adadelta}
M.~D. Zeiler.
\newblock Adadelta: An adaptive learning rate method.
\newblock \emph{arXiv preprint arXiv:1212.5701}, 2012.

\bibitem[Zeiler et~al.(2013)Zeiler, Ranzato, Monga,
  et~al.]{zeiler2013rectified}
M.~D. Zeiler, M.~Ranzato, R.~Monga, et~al.
\newblock On rectified linear units for speech processing.
\newblock In \emph{ICASSP}, 2013.

\end{thebibliography}
\bibliographystyle{abbrvnat}
}

\end{document}